\pdfoutput=1
\documentclass[11pt]{article}

\usepackage{EMNLP2022}
\usepackage{graphicx}
\usepackage{subfigure}
\usepackage{booktabs} % for professional tables
\usepackage{xspace}

\usepackage{algorithm}  
\usepackage{algpseudocode}  

\usepackage{CJK} %为了使用中文

\usepackage{multirow}  % 为了使用跨行表格

\usepackage{hyperref}

\usepackage{amsmath}
\usepackage{amssymb}
\usepackage{mathtools}
\usepackage{amsthm}

\usepackage[capitalize,noabbrev]{cleveref}

\theoremstyle{plain}

\theoremstyle{definition}

\theoremstyle{remark}

\newcommand{\proposed}{ZeroPrompt\xspace}
\newcommand{\tabincell}[2]{\begin{tabular}{@{}#1@{}}#2\end{tabular}}  

\usepackage{pifont}
\usepackage[textsize=tiny]{todonotes}

\usepackage{makecell}

\usepackage{times}
\usepackage{latexsym}

\usepackage[T1]{fontenc}
\usepackage[utf8]{inputenc}

\usepackage{microtype}

\usepackage{inconsolata}

\title{ZeroPrompt: Scaling Prompt-Based Pretraining to 1,000 Tasks Improves Zero-shot Generalization}

\author{
Hanwei Xu$^{*}$, Yujun Chen$^{*}$, Yulun Du$^{*}$, \\
{ \bf Nan Shao, Yanggang Wang, Haiyu Li, Zhilin Yang$^{\dagger}$} \\
    Recurrent AI \\
    \texttt{\{xuhanwei, chenyujun, duyulun, kimi\_yang\}@rcrai.com}
}

\begin{document}
\maketitle
\begin{abstract}
We propose a multitask pretraining approach \proposed for zero-shot generalization, focusing on task scaling and zero-shot prompting.
While previous models are trained on only a few dozen tasks, we scale to 1,000 tasks for the first time using real-world data. This leads to a crucial discovery that task scaling can be an efficient alternative to model scaling; i.e., the model size has less impact on performance with an extremely large number of tasks. Our results show that on the datasets we consider, task scaling can improve training efficiency by 30 times in FLOPs.
Empirically, \proposed substantially improves both the efficiency and the performance of zero-shot learning across a variety of academic and production datasets.
% Further study on different test tasks reveals that task distribution matters to the task scaling effect, and the performance of zero-shot generalization for disimilar test tasks does not increase monotonically with more training tasks.
% \footnote{Code and data will be made public upon acceptance.}
\end{abstract}

{\let\thefootnote\relax\footnote{{$*$ Equal contribution}}}
{\let\thefootnote\relax\footnote{{$\dagger$ Corresponding author}}}

\section{Introduction}

% Pretrained language models, such as BERT \cite{devlin-etal-2019-bert}, XLNet \cite{yang2019xlnet}, T5 \cite{JMLR:v21:20-074}, and GPT \cite{radf2019lmmultitask}, are often finetuned for downstream natural language processing tasks, which has been shown to improve performance over non-pretrained models. However, this pretraining-finetuning paradigm still relies on a relatively large set of labeled data for each downstream task to obtain competitive performance. 

% Generalizing language models to a variety of downstream tasks draws large amount of attention. 
Recent progress like GPT-3 \cite{NEURIPS2020_gpt3} demonstrates the possibility of prompting on larger-scale models for zero-shot learning, but the performance of zero-shot generalization still falls short on many tasks compared to fully-supervised finetuning.
Further, other works proposed to include a set of supervised tasks into pretraining~\cite{zhong2021adapting,wei2021finetuned,sanh2021multitask}, and prompts are often used in the framework to unify the tasks.
\citet{zhong2021adapting} converted different datasets into a unified ``yes/no'' question answering format with label descriptions. FLAN~\cite{wei2021finetuned} extended the scope by considering more task types and a larger model. T0~\cite{sanh2021multitask} collected a large set of diverse prompts for each task to further enhance performance.

% wang2022benchmarking

\begin{figure}[t]
    \centering
    \includegraphics[width=0.4\textwidth]{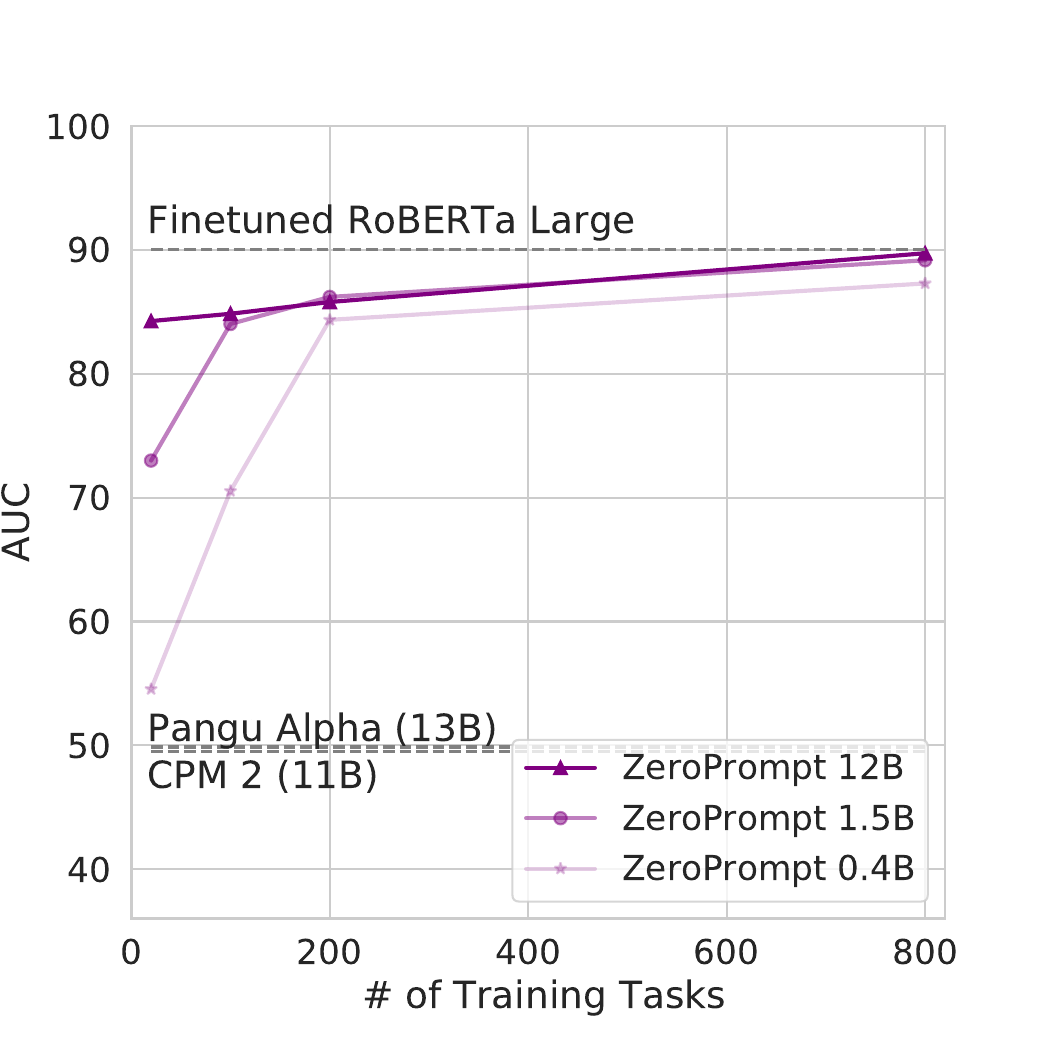}
    \caption{Task scaling vs model scaling. The horizontal axis is the number of training tasks, and the vertical axis is the zero-shot performance on unseen tasks.
    % Task scaling is an efficient alternative to model scaling.
    % With an extremely large number of training tasks, the model size has less impact on performance.
    % Moreover, task scaling consistently improves performance at various model scales.
    RoBERTa-Large was finetuned in a fully-supervised manner, while Pangu Alpha, CPM-2 and our ZeroPrompt were zero-shot prompted.
    % We report zero-shot performance without any labeled data.
}
    \label{fig:main_cmp}
\end{figure}

Despite the effects of model scaling and prompts scaling~\cite{wei2021finetuned,sanh2021multitask} have been explored, only dozens of training tasks are exploited in these works. It is still not clear how scaling the number of training tasks to hundreds even thousands of tasks affects the performance of multitask pretraining. We hypothesize that task scaling plays an important role in training generalizable zero-shot systems and explore the limits of task scaling using 1,000 tasks. Interestingly, our empirical study reveals that task scaling can be an efficient alternative to model scaling, as shown in Figure \ref{fig:main_cmp}.
With an extremely large number of training tasks, the model size has less impact on performance.
A 0.4B model can achieve comparable zero-shot performance to that of a 12B model, improving training efficiency by 30 times in terms of FLOPs and the serving efficiency as well. 

Our contributions can be summarized as follows. 
\begin{itemize}
    \item We scale the number of tasks to 1,000 in multitask pretraining for the first time. Our study reveals a crucial finding that on the datasets we consider, task scaling is an efficient alternative to model scaling.
    \item Our experiments demonstrate that task scaling improves both the efficiency and the performance of zero-shot learning.
\end{itemize}
\section{Related Work}

Pretrained language models, like BERT~\cite{devlin-etal-2019-bert}, XLNet~\cite{yang2019xlnet}, T5~\cite{JMLR:v21:20-074} and GPTs~\cite{NEURIPS2020_gpt3,radf2019lmmultitask}, have achieved strong performance on various NLP tasks. In some cases, pretrained models can perform well with only a few training samples~\cite{liu2021gpt,schick-schutze-2021-exploiting}, or even without any training sample~\cite{Shen2021ConterfactualGZ,sanh2021multitask}. 
% These works mainly focus on unsupervised pretraining on a large-scale corpus.

It has been shown that augmenting unsupervised pretraining with supervised data can significantly improve task performance during finetuning~\cite{chen2020big, gururangan2020don}.
Some recent studies followed this idea and obtained improved few-shot or zero-shot generalization in the same manner.
For instance, Mishra et al.~\cite{swaroopmishra2021crosstask} built a dataset with task instructions, and
CROSSFIT~\cite{ye2021crossfit} introduced a repository of few-shot text-to-text tasks.
FLAN~\cite{wei2021finetuned} and T0~\cite{sanh2021multitask} applied instruction-tuning of many tasks with 137B and 11B parameters, respectively. 
ExT5~\cite{2021ext5} applies multitask pretraining as well, but it focuses on multitask cotraining transfer instead of zero-shot generalization.
% to a large-scale decoder model with 137B parameters, while T0~\cite{sanh2021multitask} trained an encoder-decoder model with 11B parameters on a large number of professionally crowd-sourced prompts.
Our \proposed utilizes labeled data in the pretraining phase, and we aim at studying the task scaling law of zero-shot generalization by adopting 1,000 real-world tasks.

% \subsection{Domain Generalization}
% Domain generalization addresses the model's ability to generalize to unseen test domains with data from several different but related domains~\cite{Wang2021GeneralizingTU}.
% Previous approaches in domain generalization are mainly based on gradient operations~\cite{gulrajani2020search,arjovsky2019invariant,mansilla2021domain,kim2021selfreg} or disentanglement learning~\cite{wang2020cross,peng2020domain2vec,peng2019domain}. 
% Different from the above problem formulation, we focus on zero-shot task generalization to unseen tasks with the benefit of prompting to unify different NLP tasks by having the same data format.
\section{\proposed}
We follow the same framework of multitask zero-shot learning in ~\cite{wei2021finetuned,sanh2021multitask}, where models are pretrained on a variety of tasks and then tested on held-out unseen tasks.

\begin{table}[!t]
\small
\centering
\begin{tabular}{lc}
\toprule
\textbf{Task type} & \textbf{\# of Tasks}  \\
\midrule
\scriptsize Sentiment Analysis (\textbf{SENTI})                 & 17 (4,13)          \\ 
\scriptsize News Classification (\textbf{NEWS})                 & 9  (4,5)          \\
\scriptsize Intent Classification  (\textbf{INTENT})            & 4  (1,3)      \\
\scriptsize Natural Language Inference. (\textbf{NLI})          & 2  (1,1)   \\ 
\scriptsize Sentence Similarity.  (\textbf{STS})                & 13 (3,10)  \\ 
\scriptsize Paraphrase (\textbf{PARA})                          & 1  (0,1)     \\ 
\scriptsize Question Answer Matching.  (\textbf{QAM})           & 1  (0,1)   \\
\scriptsize Machine Reading Comprehension (\textbf{MRC})        & 10 (5,5)    \\ 
\scriptsize Name Entity Recognition   (\textbf{NER})            & 9  (3,6)\\
\scriptsize Summarization  (\textbf{SUMM})                       & 9  (3,6)\\
\scriptsize Keywords       (\textbf{KEYS})                      & 3  (0,3) \\ 
\scriptsize Winograd Schema Challenge  (\textbf{WSC})           & 1  (0,1) \\ 
\scriptsize App Classification    (\textbf{APP})                & 1  (0,1) \\ 
\scriptsize Production tasks (\textbf{Objection})                                    & 110 (85,25) \\
\scriptsize Production tasks (\textbf{Profile})                                     & 345 (268,77) \\
\scriptsize Production tasks (\textbf{Execution})                                     & 310 (240,70) \\
\scriptsize Production tasks (\textbf{Mention})                                     & 125 (97,28) \\
\scriptsize Production tasks (\textbf{Violation})                                     & 90 (70,20) \\
\scriptsize Production tasks (\textbf{Acception})                                     & 50 (38,12) \\
\midrule
In total                                            &  1110 (824,286) \\
\bottomrule
\end{tabular}
\caption{The number of tasks for each task type. Numbers in brackets stand for the number of tasks for training and testing, respectively. e.g. SENTI has 4 tasks for training and 13 for testing.}
\label{tab:datasets_used_count}
\end{table}

% \begin{figure*}[!t]
%     \centering
%     \includegraphics[width=0.8\textwidth]{figures/paradigm.pdf}
%     \caption{Different NLP paradigms. (a) Traditional pretraining-finetuning paradigm, requiring considerable labeled data for finetuning (b) \proposed pipeline, only need a tiny validation set for prompt search.}
%     \label{fig:compare_figure}
% \end{figure*}

% \subsubsection{Datasets}
\subsection{Datasets for Scaling to 1,000+ Tasks}\label{subsection:Datasets}
We collected 80 public Chinese NLP tasks and further
% the Chinese Language Understanding Evaluation benchmark (CLUE),
% the Chinese Natural Language Processing Conference Competition (NLPCC),
% and many other well-known Chinese NLP competitions.
% Considering that the number of open-source datasets is still insufficient for studying task scaling,
acquired over 1,000 real-world datasets from our production systems to investigate the task number scaling law. The number of tasks in each task type is listed in Table~\ref{tab:datasets_used_count}, where we define task types following previous work and intuitive knowledge.
The task taxonomy of the production datasets is presented in Appendix~\ref{appendix:datasets}, consisting of 6 task types from 10 different domains.

We split the public datasets and the production datasets into training tasks and testing tasks, as shown in Table~\ref{tab:datasets_used_count}.
Different from FLAN~\cite{sanh2021multitask} or T0~\cite{wei2021finetuned}, our test set contains a more diverse set of task clusters.
% , and the number of test datasets is twice that of train sets for proper evaluation.
% We argue that this is the practical real-world setting with a large number of different unseen tasks in the test set.
Detailed train/test splits can be found in Table~\ref{tab:summary_datasets}.
To simulate real-world NLP production systems at scale, where the costs for data labeling are expensive, we sample 128 examples per class for each classification task and 256 examples for each generation task to build the training set\footnote{Only 512 data points are sampled for the iflytek dataset as it has over 100 classes}.

\subsection{Prompt Design}
Although large-scale pretrained models with prompting show promising results on zero-shot generalization to unseen tasks without any labeled data, prompt design is of vital importance to their performance. We applies both the hard prompt, which is composed of label candidates and task descriptions, and the soft prompt at the mulitask pretraining stage, details of prompt design can be found in Appendix~\ref{appendix:prompt_design}.

\section{Experiments} \label{sec:exp}
\begin{table*}[t]
	\small
    \centering
    \begin{tabular}{cccccccc}
    \toprule
%    \hline\hline
         task type & task & \textbf{CPM-2} & \textbf{Pangu-$\alpha$} & \textbf{T5} & \textbf{RoBERTa} & \textbf{\proposed} & \textbf{T5} \\ 
         ~ & ~ & Zero-Shot & Zero-Shot & Zero-Shot & Finetuning & Zero-Shot & Finetuning \\
         \midrule
\multirow{3}{*}{\textbf{SENTI}}  & \scriptsize online\_shopping\_10cats & 80.60 & 61.99 & 71.88 & $95.30_{(0.42)}$ & $\textcolor{blue}{95.90}_{(0.24)}$ & $96.94_{(0.26)}$ \\ 
                              ~ & \scriptsize nlpcc2014\_task2         & 68.53 & 56.22 & 60.06 & $72.09_{(0.80)}$ & $\textcolor{blue}{80.49}_{(0.80)}$ & $80.67_{(0.21)}$ \\
                              ~ & \scriptsize SMP2019\_ECISA           & 29.04 & \textbf{40.41} & 31.21 & $69.45_{(1.65)}$ & $38.46_{(0.33)}$ & $74.15_{(0.30)}$ \\ \midrule
\multirow{1}{*}{\textbf{NEWS}}   & \scriptsize CCFBDCI2020              & 49.57 & 38.09 & 27.48 & $90.73_{(0.58)}$ & $\textbf{80.50}_{(1.68)}$ & $96.53_{(0.41)}$ \\ \midrule
\multirow{1}{*}{\textbf{INTENT}} & \scriptsize catslu\_traindev         & 62.63 & 46.65 & 11.27 & $91.09_{(2.33)}$ & $\textbf{90.48}_{(0.78)}$ & $94.42_{(0.66)}$ \\ \midrule
\multirow{1}{*}{\textbf{NLI}}    & \scriptsize ocnli\_public            & 33.76 & 38.58 & 30.51 & $54.70_{(0.53)}$ & $\textbf{46.16}_{(1.87)}$ & $58.15_{(1.61)}$ \\ \midrule
\multirow{2}{*}{\textbf{STS}}    & \scriptsize CBLUE-CHIP-STS           & 44.15 & 56.40 & 44.94 & $80.28_{(1.08)}$ & $\textbf{77.90}_{(0.59)}$ & $82.45_{(2.07)}$ \\
                              ~ & \scriptsize sohu-sts-B-ss            & 33.50 & 54.94 & 43.46 & $89.71_{(0.68)}$ & $\textbf{79.85}_{(1.03)}$ & $89.85_{(0.86)}$ \\ \midrule
\multirow{1}{*}{\textbf{QAM}}    & \scriptsize nlpcc2016-dbqa           & 49.90 & 56.08 & 51.69 & $56.31_{(1.51)}$ & $\textcolor{blue}{62.61}_{(3.64)}$ & $76.76_{(1.95)}$ \\ \midrule
\multirow{1}{*}{\textbf{PARA}}   & \scriptsize PAWS-X                   & 48.08 & 53.06 & 48.08 & $53.51_{(0.53)}$ & $\textcolor{blue}{54.90}_{(0.37)}$ & $59.04_{(0.51)}$ \\ \midrule
\multirow{1}{*}{\textbf{MRC}}    & \scriptsize cmrc2018\_public         & 8.51  & 11.61 & 5.94  & -                & $\textbf{35.50}_{(0.73)}$ & $61.00_{(0.80)}$ \\ \midrule
\multirow{2}{*}{\textbf{NER}}    & \scriptsize msra\_ner                & 3.11  & 9.81* & 21.44 & -                & $\textbf{58.17}_{(4.40)}$ & $65.37_{(2.65)}$ \\
                              ~ & \scriptsize CMeEE                   & 1.18     & 9.44* & 6.77  & -                & $\textbf{24.84}_{(0.94)}$ & $29.34_{(2.84)}$ \\ \midrule
\multirow{1}{*}{\textbf{SUMM}}    & \scriptsize EDU\_SUMM                & 1.05  & 10.02 & 2.21  & -                & $\textbf{14.80}_{(3.15)}$ & $16.97_{(2.11)}$ \\ \midrule
\multirow{1}{*}{\textbf{KEYS}}   & \scriptsize COTE-MFW                 & 1.29  & 4.91  & 7.05  & -                & $\textbf{50.34}_{(9.01)}$ & $79.35_{(1.08)}$ \\ \midrule
\multirow{1}{*}{\textbf{WSC}}    & \scriptsize cluewsc2020\_public      & \textbf{57.74} & 44.93 & 44.08 & $71.99_{(3.32)}$ & $47.98_{(4.18)}$ & $72.81_{(2.19)}$ \\ \midrule
\multirow{1}{*}{\textbf{APP}}  & \scriptsize iflytek\_public          & 4.77  & 7.85  & 1.69  & $50.34_{(0.61)}$ & $\textbf{26.14}_{(1.02)}$ & $53.33_{(1.05)}$ \\ \midrule
\multirow{10}{*}{\textbf{Production}} 
                                 & \scriptsize Return Commitment             & 36.28 & 51.83 & 53.28 & $96.16_{(0.21)}$ & $\textbf{95.53}_{(0.24)}$ & $96.78_{(0.62)}$\\   
                              ~ & \scriptsize Heating Supply               & 44.89 & 31.61 & 44.57 & $97.48_{(0.30)}$ & $\textcolor{blue}{99.22}_{(0.35)}$ & $98.91_{(0.59)}$\\
                              ~ & \scriptsize Return Amount                & 53.26 & 46.09 & 55.90 & $90.71_{(0.33)}$ & $\textbf{89.48}_{(0.56)}$ & $90.86_{(0.47)}$\\
                              ~ & \scriptsize Registration Discount             & 55.09 & 50.34 & 56.25 & $88.68_{(0.40)}$ & $\textbf{88.48}_{(0.51)}$ & $89.88_{(0.65)}$\\
                              ~ & \scriptsize Operation Guidance                & 57.97 & 47.71 & 54.52 & $90.78_{(0.35)}$ & $\textbf{78.24}_{(1.41)}$ & $92.80_{(0.84)}$\\
                              ~ & \scriptsize Promise for Refunding         & 46.80 & 49.35 & 48.57 & $93.71_{(0.24)}$ & $\textcolor{blue}{94.28}_{(0.56)}$ & $91.40_{(1.13)}$\\
                              ~ & \scriptsize Households Heating Plant      & 63.37 & 69.66 & 48.71 & $96.59_{(0.47)}$ & $\textcolor{blue}{98.22}_{(0.52)}$ & $97.39_{(0.59)}$\\
                              ~ & \scriptsize Refunding Amount              & 48.48 & 52.58 & 49.67 & $83.78_{(0.52)}$ & $\textcolor{blue}{88.03}_{(0.83)}$ & $83.74_{(1.67)}$\\
                              ~ & \scriptsize Cost Abatement                & 43.18 & 48.13 & 51.51 & $80.30_{(0.92)}$ & $\textcolor{blue}{81.88}_{(0.22)}$ & $81.40_{(1.02)}$\\
                              ~ & \scriptsize WeChat Operation              & 45.45 & 51.37 & 47.79 & $82.28_{(0.59)}$ & $\textbf{78.25}_{(0.26)}$ & $83.53_{(1.59)}$\\ \midrule
\multicolumn{2}{c}{\textbf{AVG}}                                     & 39.71 & 40.73 & 37.80 & -                & $\textbf{68.76}_{(1.48)}$ & $77.55_{(1.14)}$ \\
\multicolumn{2}{c}{\textbf{AVG excl. GEN}}                           & 48.05 & 47.90 & 44.42 & $80.73_{(0.85)}$ & $\textbf{76.04}_{(1.02)}$ & $83.72_{(0.94)}$ \\
\bottomrule
    \end{tabular}
    \caption{Main results of \proposed (1.5B) and other zero-shot/finetuning baselines. The numbers in brackets are the standard deviations of results with 5 different random seeds.
    -: We do not finetune RoBERTa on generation tasks because it is an encoder-only model.
    *: Only part of the test set is sampled for evaluation due to the computation burden. \textcolor{blue}{Blue} numbers indicate the cases where \proposed scores better than finetuned RoBERTa and \textbf{bold} numbers indicate the cases where \proposed achieves the best zero-shot performance. }
    \label{tab:main_results}
\end{table*}

\subsection{Experiment Setups}

% \subsubsection{Models}
We compare \proposed with state-of-the-art large-scale Chinese pretrained models, Pangu-$\alpha$ (13B decoder)~\cite{zeng2021pangualpha}, CPM-2 (11B encoder-decoder)~\cite{Zhang2021cpm2}, and the finetuned RoBERTa-large model~\cite{liu2019roberta}. All finetuned baselines were trained one task at a time. We use a encoder-decoder model and apply both unsupervised pretraining and multitask prompted supervised pretraining. Training details of \proposed can be found in Appendix~\ref{appendix:training_detail}.

% We repeat experiments, including multitask pretraining and finetuning of RoBERTa, T5, five times with different random seeds to reduce variance.
% Training and test datasets details are presented in Sec~\ref{subsection:Datasets}.
% Details regarding evaluation metrics can be seen in Appendix~\ref{subsec:appendix_metric}.

% \begin{figure*}[t]
%     \centering
%     \includegraphics[width=0.40\textwidth]{figures/task_scale_cross_domain.pdf}
%     \includegraphics[width=0.40\textwidth]{figures/task_scale_cross_type.pdf}
%     \caption{Out-of-domain (left) and cross-task-type (right) zero-shot performance with different number of training tasks. Few-shot validation is \textbf{not} employed to exclude potential confounding factors from the study.}
%     \label{fig:task_scale_cross_domain_task_type}
% \end{figure*}

\subsection{Main Results}
\subsubsection{Power of Task Scaling} 
% \zy{Should this be put into Experiments? This should be the main results I guess.}
\label{subs: scale_experiment}

% We explore the limits of zero-shot performance of multitask prompted pretraining, using over 1,000 tasks from 10 domains. 
% Since model scaling has been shown to have significant impact on zero-shot performance~\cite{NEURIPS2020_gpt3}, we also provide the results of three models with 0.4B, 1.5B and 12B parameters.
To study the law of task scaling, we trained \proposed on a mixture of public data and production data, and increased the number of production training tasks from 20 to 800.
Zero-shot performance on unseen production test tasks are presented in Figure~\ref{fig:main_cmp}.
Larger models have much better zero-shot performance with a limited number of training tasks.
However, the performance gains from larger models decrease when more training tasks are added. 
% The 0.4B model achieves a score of 87 with 800 training tasks, which scores better than the 12B model with only 200 training tasks. 
Generally, if we scale the number of training tasks, small models can still achieve impressive zero-shot performance, substantially improving training efficiency by 30 times in FLOPs (0.4B vs 12B) as well as the serving efficiency.

\begin{figure}[t]
    \centering
    \includegraphics[width=0.40\textwidth]{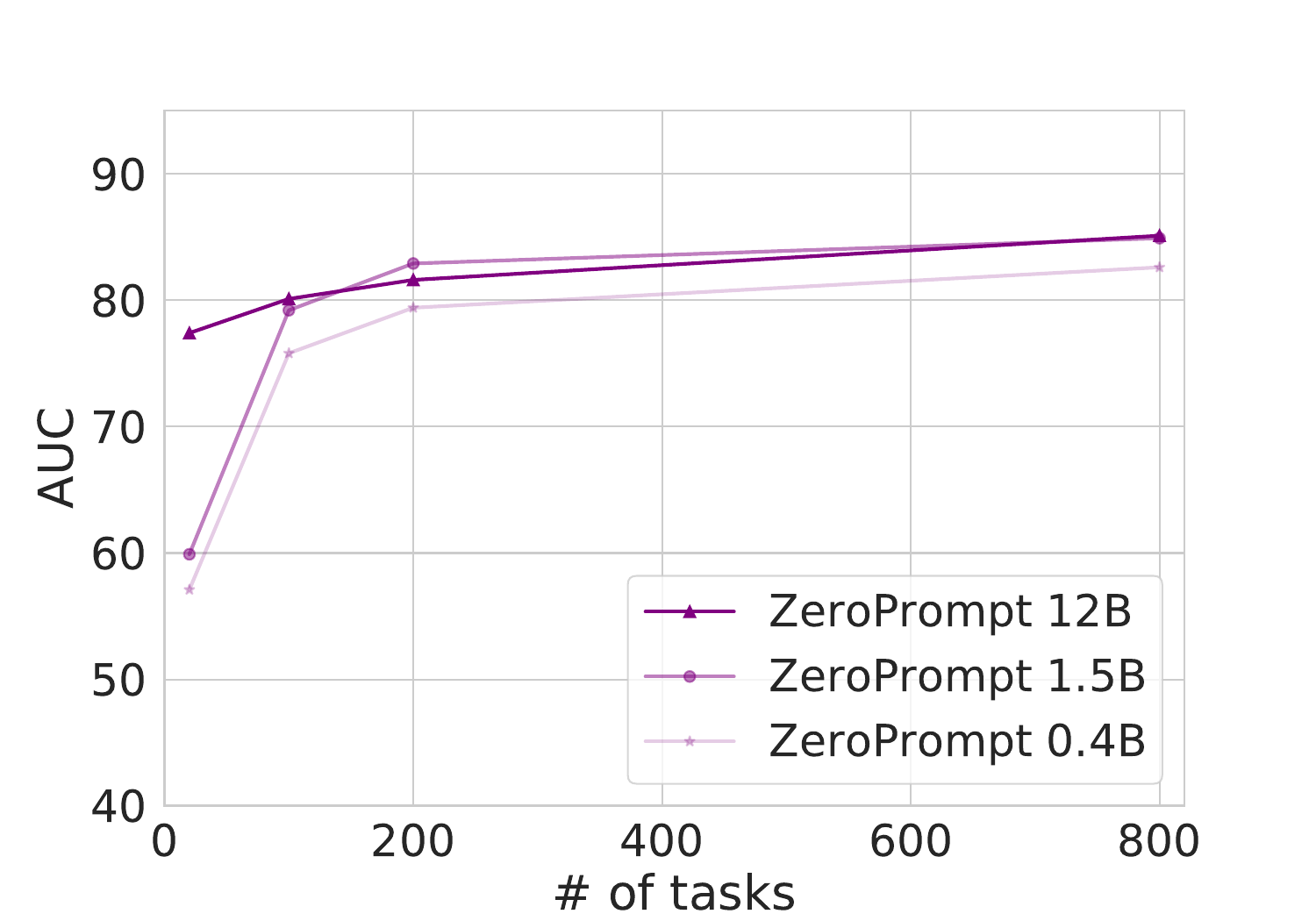}
    \caption{Zero-shot performance on cross-task-type tasks with different number of training tasks. }
    \label{fig:task_scale_cross_task_type}
\end{figure}

\subsubsection{Comparison with Other Baselines}

% In this section, we compare \proposed with other strong zero-shot and fully-supervised baselines on unseen testing datasets.
% Note that due to limited space, only part of the reserved testing tasks, specifically 17 academic and 10 production datasets, are included in Table~\ref{tab:main_results} for comparison.

Results on the reserved testing tasks are shown in Table~\ref{tab:main_results}, in the zero-shot setting, \proposed significantly improves the performance of T5 from 37.80 to 68.76 with a boost of 30.96 points, outperforming previous PTMs, CPM-2 and Pangu-$\alpha$, by a large margin of 28 points.
Notably, \proposed is comparable to or even better than a finetuned RoBERTa-large model on some academic and production datasets.
Compared to the overall score of the finetuned RoBERTa, \proposed is only 4.7 points short. This is quite ecstatic considering that \proposed did not use any labeled data for tuning. 
% The finetuned T5 is better than RoBERTa, and the gap between \proposed and the finetuned T5 is less than 8 points.

\subsection{Discussions}
\subsubsection{Task Scaling vs Sample Scaling}

While task scaling by definition also increases the number of training samples, we also decouple the effects of task scaling and sample scaling in Table \ref{tab:task_num_data_num}. The numbers of total samples are the same for ``80 tasks with 1280 shots'' and ``800 tasks with 128 shots'', but the latter shows considerably better performance---4.8 and 3.0 points improvement for the 0.4B model and the 1.5B model, respectively.
% Moreover, task scaling closes the gap between different model sizes; this effect is less prominent for sample scaling.

% We also study the efficiency of task scaling and sample scaling, and the results are given in Table~\ref{tab:task_num_data_num}. Although the number of data points for 100 tasks with 1280-shot is the same as 800 tasks with 128-shot, the latter shows considerable better performance by 4.8, 3.0 and - points for models with 0.4B, 1.5B and 12B parameters, respectively. The empirical results show that task scaling is more efficient than sample scaling to boost the zero-shot performance, especially for small and medium-sized models.

\subsubsection{Unsupervised Data vs Supervised Data}
\begin{table}[t]
\footnotesize
    \centering
    \begin{tabular}{cccc}
    \toprule
    Model size & \makecell[c]{100 tasks \\ 128-shot} & \makecell[c]{80 tasks \\ 1280-shot} & \makecell[c]{800 tasks \\ 128-shot} \\
    \midrule
    0.4B & 70.5 & 82.5 & 87.3 \\
    1.5B & 84.0 & 86.2 & 89.2 \\
    12B  & 84.8 & 88.7 & 89.4 \\
    \bottomrule
    \end{tabular}
    \caption{Task scaling vs sample scaling.}
    \label{tab:task_num_data_num}
\end{table}

\begin{table}[h]
\centering
\footnotesize
\begin{tabular}{cccc}
\toprule
    Model & 0.4B & 1.5B & 12B \\ \midrule
    LM loss & 1.9 & 1.7 & 1.5 \\
    Sup loss & 0.19 & 0.17 & 0.19 \\
\bottomrule
\end{tabular}
\caption{Language modeling (LM) and supervised (Sup) validation loss of models with different sizes.}
\label{tab:loss}
\end{table}

Zero-shot performance is attributed to both supervised tasks and the LM task. As we increase the number of supervised tasks, they outweigh the LM task. Meanwhile, these supervised tasks have much less data to fit than the LM task, which makes smaller models viable choices.
Table \ref{tab:loss} shows that smaller models have similar losses on supervised tasks but higher losses on LM, compared to larger models. This explains why task scaling can be an alternative to model scaling.

\subsubsection{Effect of Task Distribution}
\begin{figure}[t]
    \centering
    \includegraphics[width=0.40\textwidth]{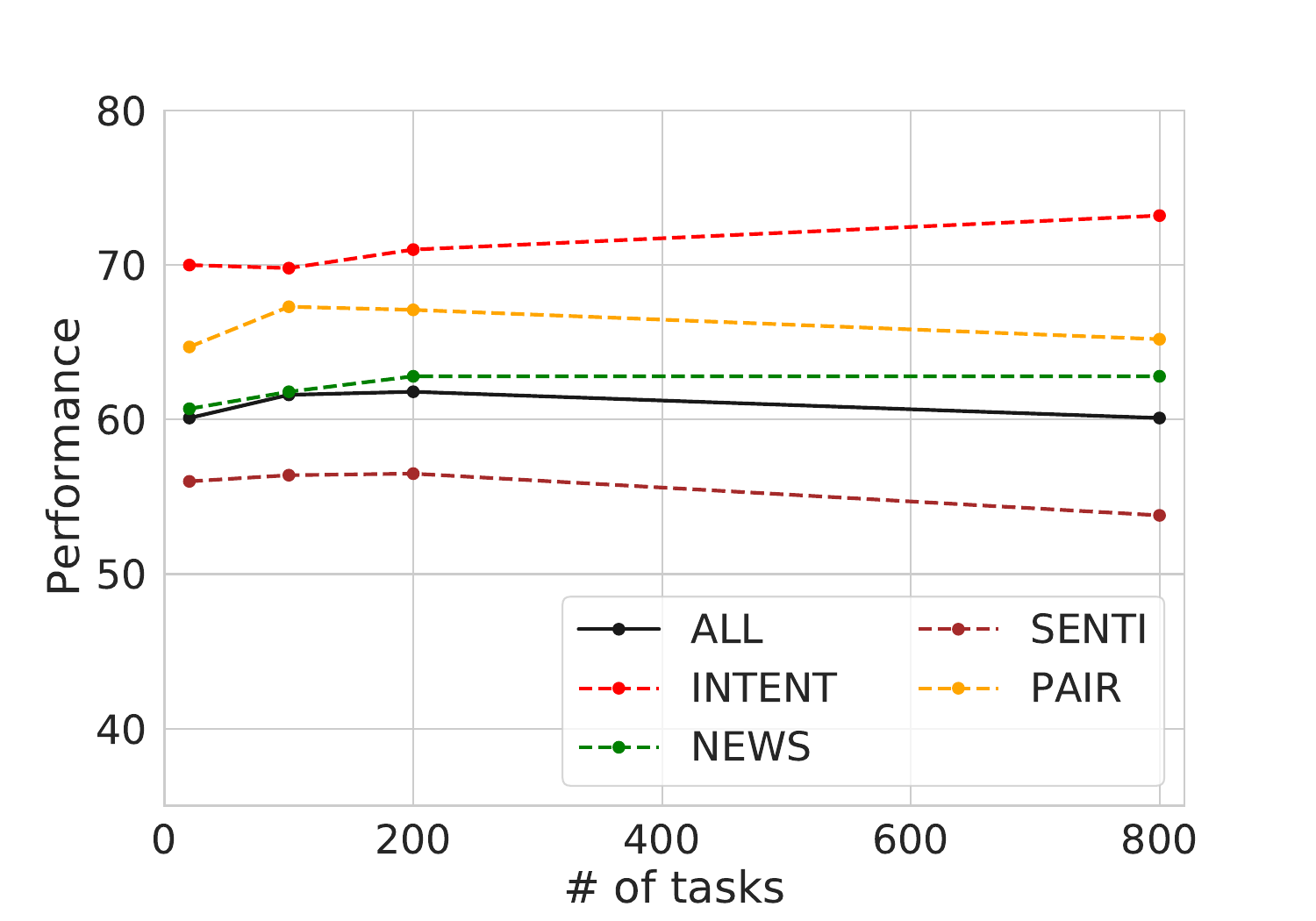}
    \caption{Zero-shot performance of 1.5B model on public datasets with different number of production training tasks.}
    \label{fig:task_scale_general_test_1b}
\end{figure}

To validate the zero-shot performance on cross-task-type tasks, we select production tasks from two task types for testing and the rest for training, as presented in Figure~\ref{fig:task_scale_cross_task_type}. It can be seen that task scaling still leads to significant improvement of zero-shot performance on cross-task-type tasks. On the other hand, Figure~\ref{fig:task_scale_general_test_1b} shows the zero-shot performance on public datasets. For some tasks like INTENT, the scaling of production tasks is helpful, but the result could be different for other tasks like SENTI. The average performance of all public datasets does not increase monotonically with more training tasks. We suppose the reason is that the task distribution of production data is different from that of public tasks. Therefore, only part of public tasks benefit from the scaling of production training tasks. We also study the effect of cross task type transfer on public tasks, the results can be found in Appendix~\ref{appendix:cross_type_transfer}.

\section{Conclusions}
In this paper, we propose \proposed, a multitask prompted pretraining method that significantly improves the zero-shot generalization ability of language models. In our experiments, we collect over 1,000 real-world production tasks to study the task scaling law. We find that on our considered datasets, the zero-shot performance gap between small and large models becomes less significant when having more training tasks. As a result, task scaling can substantially improve training and serving efficiency.

\section{Limitations}
% We show that task scaling, which could now be seen as an alternative to model scaling, improves both the efficiency and performance of zero-shot learning. 
Our results regarding the effect of task scaling on zero-shot performance still have a few limitations.
% and it is possible that zero-shot performance could be further improved by studying those problems in the future.
Specifically, We control our study by only increasing the number of tasks collected from our production system, and they might only represent a subset of all the NLP problems. In addition, for different testing tasks in the public datasets, the zero-shot performance might not increase with the scaling of production training tasks. Therefore, the conclusion that task scaling can significantly boost zero-shot performance is limited to the case where training and test tasks share some similarity in distribution, but not a general conclusion for arbitrary distributions.
It also remains an open problem as how to quantitatively characterize the distribution similarity between training and test tasks.
% stands on our collected production tasks, but not a general conclusion.
% The problem of how to choose a better training task distribution is left for future work.
% , since it is not relevant to our main focus.
% 2) Our tasks are not exhaustive, because our real-world production data might only represent a subset of all the NLP problems.
% However, it is challenging to collect more diverse data for the study of task scaling, because publicly available data is often limited in terms of the number of tasks. Therefore, we choose to use production data to initiate such a study.
% 3) The conclusion that task scaling can significantly boost zero-shot performance stands on our collected production datasets, but not a general conclusion for all NLP tasks.
We hope our results could encourage future work on addressing these limitations to further explore the potential of zero-shot learning.
% We will also publish related data to facilitate research in this direction.

\bibliography{custom}
\bibliographystyle{acl_natbib}

% \bibliography{custom}
% \bibliographystyle{acl_natbib}

\newpage
\appendix
\cleardoublepage
\section{Appendix}
\label{sec:appendix}

\subsection{Datasets}
\label{appendix:datasets}
For fair evaluation of zero-shot generalization, 
we investigate and collect diverse public Chinese NLP datasets with different task types.
% In the following, we provide details of each task cluster and the data preprocessing approach to avoid test set contamination.
The summary of all datasets used in the experiments is presented in Table~\ref{tab:summary_datasets}, including train/test task split and metrics of each task. In total, we have 13 task types of public datasets and 6 task types of production datasets.
% In the following, we briefly introduce each task type, in case some readers are not familiar with these task types.

\subsubsection{Public Datasets}
\begin{itemize}
	\item \textbf{Sentiment Analysis} requires the model to determine whether the sentiment of a piece of text is positive or negative.
	\item \textbf{News Classification} asks the model to predict the topic of a news article.
	\item \textbf{Intent Classification} asks the model to predict the intent of a person given one of his/her words.
	\item \textbf{Machine Reading Comprehension Question Answering} requires the model to answer a question given a document where the answer can be derived.
	\item \textbf{Natural Language Inference} asks the model to tell the relation of two sentences is neutral, entailment or contradiction.
	\item \textbf{Sentence Similarity} asks the model to predict whether two sentences are similar or not.
	\item \textbf{Paraphrase} asks the model to tell whether two sentences with much lexical overlap are semantically equivalent.  
	\item \textbf{Question Answer Matching} asks the model to reason whether the given two sentences can form a valid question answering pair. 
	\item \textbf{Name Entity Recognition} requires the model to find all entities in the given piece of text.
	\item \textbf{Summarization} requires the model to give a summary with one or two sentences of the given long document.
	\item \textbf{Keywords} asks the model to extract keywords from the given sentence.
	\item \textbf{Winograd Schema Challenge}, the sample of which composes a sentence, a pronoun and an entity in the sentence, requires the model to tell whether the pronoun refers to the entity.
	\item \textbf{App Classification} asks the model to tell which type of App the given introduction is about, and there are hundreds of target App categories.
\end{itemize}

\subsubsection{Production Datasets}

\begin{figure}[t]
    \centering
    \includegraphics[width=0.45\textwidth]{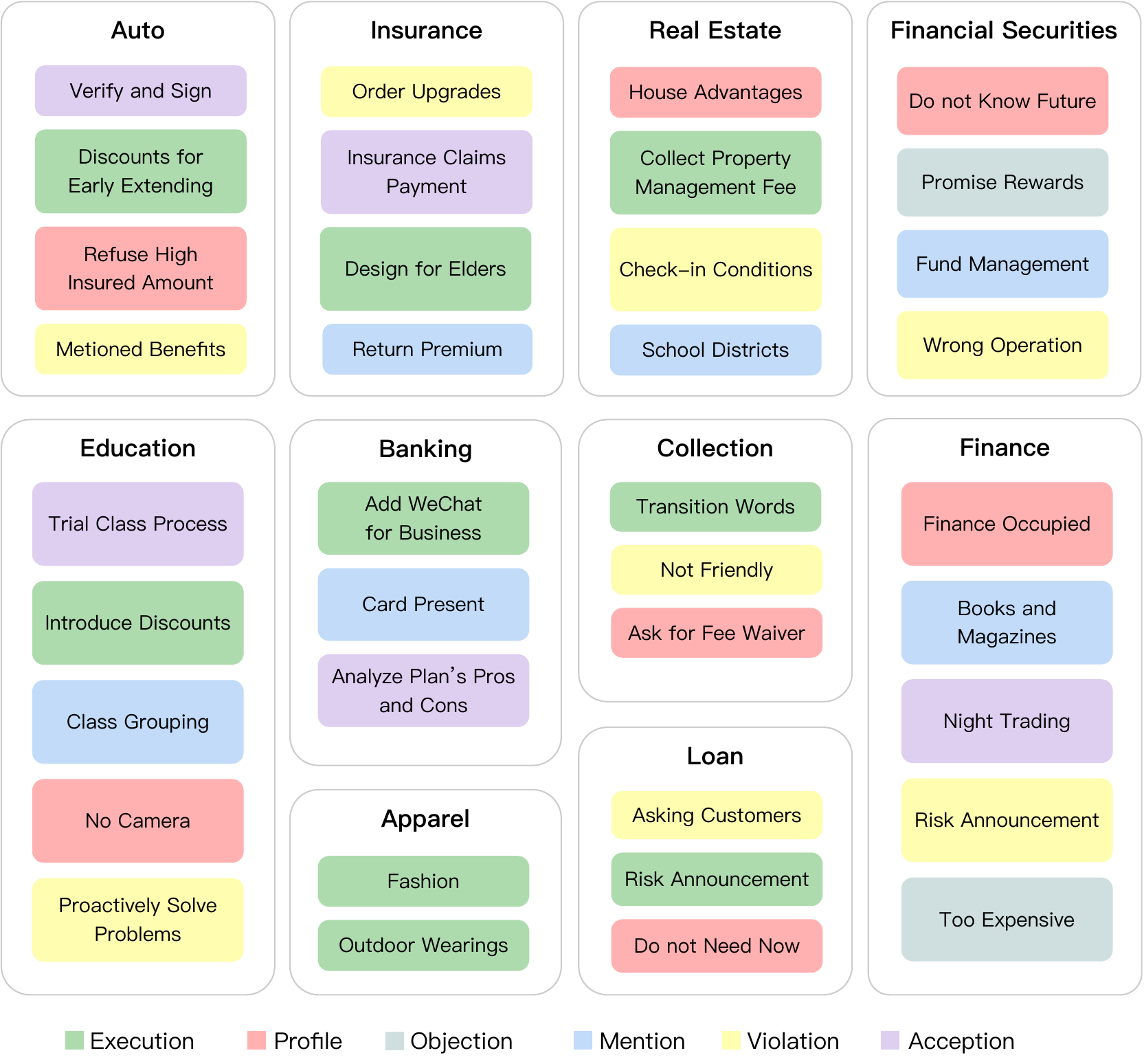}
    \caption{The task taxonomy of the real-world production datasets. The tasks are collected from commercial sales conversations in ten domains, e.g. \emph{Auto} and \emph{Insurance}. Task types are marked by different colors. For example, ``Profile'' is to predict an aspect of customer profile from a given transcribed text, and ``Acception'' is to judge whether a salesperson follows a certain sales script.}
    \label{fig:task_taxonomy}
\end{figure}

The task taxonomy of the production datasets is presented in Figure~\ref{fig:task_taxonomy}, consisting of 6 task types from 10 different domains.
As illustrated in Figure \ref{fig:task_taxonomy}, the task taxonomy of our production contains six types of natural language understanding tasks. 
We provide detailed explanation here and several examples in Table~\ref{tab:production}. 
% shows the task taxonomy and in Table \ref{tab:production}, we give several examples that sampled from these datasets.
\begin{itemize}
    \item \textbf{Objection} are datasets that we gathered from production scenario. Objection tasks are language understanding tasks where model will have to analyze whether the speaker is proposing an argument in opposition of the previous contents. 
	\item \textbf{Profile} are datasets that we gathered from realistic industrial scenario. Profile tasks are language understanding tasks similar to intent classification, where model will have to tell whether the current sentence is describing certain intention. 
% 	binary classification but with diverse definitions. \item \textbf{Execution}
	\item \textbf{Mention} are also datasets that we gathered from realistic industrial scenario. Mention tasks are language understanding tasks that model have to judge whether given sentence mentioned sales keywords. 
	\item \textbf{Violation} are also datasets that we gathered from realistic industrial scenario. Violation tasks are language understanding tasks that model will have to tell if speaker violates the sales guidelines. 
	\item \textbf{Acception} are also datasets that we gathered from realistic industrial scenario. Acception tasks are language understanding tasks that let model tell if the speaker follows systems instruction and tell sales keywords to customer. 
	\item \textbf{Execution} are also datasets that we gathered from realistic industrial scenario. Execution tasks are language understanding tasks that model will have to find out whether a salesman follow the predefined sales guidance when talking to customer. 
\end{itemize}	

% \subsubsection{English Datasets}
% We use the same test datasets as T0~\cite{sanh2021multitask}, including ANLI~\cite{nie-etal-2020-adversarial}, SuperGLUE (CB, RTE, WSC, COPA, WiC)~\cite{wang2020superglue},  Winogrande~\cite{sakaguchi2019winogrande}, HellaSwag~\cite{zellers2019hellaswag}.

\subsubsection{Avoid Test Set Contamination}
Although we split datasets into training and testing, there is non-negligible overlap between some of the training datasets and the test set. To avoid test set contamination, we follow the filter method given in \cite{NEURIPS2020_gpt3}. Specifically, we directly remove all examples in the training phase that have a 30-gram overlap with any example in the test phase.

% \newpage

% \newpage
% \balance

% \cleardoublepage

\subsection{Metric}
\label{subsec:appendix_metric}
Metrics used for diverse NLP tasks in this paper are presented in the following.

\textbf{AUC} is the abbreviation of Area Under ROC Curve. Typically, the value of AUC is between 0.5 and 1.0.

\textbf{ROUGE} is the abbreviation of Recall-Oriented Understudy for Gisting Evaluation, which is an evaluation method oriented to the recall rate of n-grams.
We use ROUGE-1 in the paper.

\textbf{Micro-F1} is used to evaluate multi-label classification tasks. It is the harmonic average of the averaged precision and recall of all labels.

\textbf{F1} measures the overlap between the prediction and the ground truth, which is typically used in span prediction tasks.

\textbf{Pos-F1} is customized for NER tasks with a text-to-text form as shown in Table~\ref{tab:appendix_NER}. It is the averaged string F1 score for positive samples, of which the true label is not "blank".
% {\begin{CJK}{UTF8}{gbsn}\shortstack{没有} \end{CJK}}

% \newpage

% \newpage
% \balance
% \cleardoublepage

\subsection{Training Details}
\label{appendix:training_detail}
In the unsupervised pretraining stage, our base T5 model is pretrained for 100k steps on a 300G web-crawled Chinese corpus with the batch size of 4096 and the sequence length of 512. 
In the multitask prompted training stage, \proposed is trained with an Adam Optimizer for 1500 more steps with a batch size of 64 and a learning rate of 3.5e-5. We repeat experiments, including multitask pretraining and finetuning of RoBERTa, T5, five times with different random seeds to reduce variance.

% \subsection{Auxiliary MLM loss}
At the stage of unsupervised pretraining, we apply the span corruption objective, a variant of Masked Language Modeling (MLM), following T5~\cite{JMLR:v21:20-074}. Meanwhile,
we also add MLM as an auxiliary loss to overcome catastrophic forgetting in the multitask pretraining phase.
\begin{equation}
    \centering
    \mathcal{L} = \lambda\cdot\mathcal{L}_{sup} + \mathcal{L}_{\text{MLM}}
    \label{eq:mlmloss}
\end{equation}

The multitask pretraining loss is given in Equation~\ref{eq:mlmloss}, where $\mathcal{L}$ is the overall training loss, $\mathcal{L}_{sup}$ is the multitask supervised loss, $\mathcal{L}_\text{MLM}$ is the MLM loss and $\lambda$ is the loss weight.
According to Table~\ref{tab:Detail_Table_for_Ablation_Studies}, \proposed obtains 1.3 point gains by adding the MLM loss, proving our suppose to avoid catastrophic forgetting.

\subsection{Prompt Design}
\label{appendix:prompt_design}

\begin{figure*}[t]
    \centering
    \includegraphics[width=1.0\textwidth]{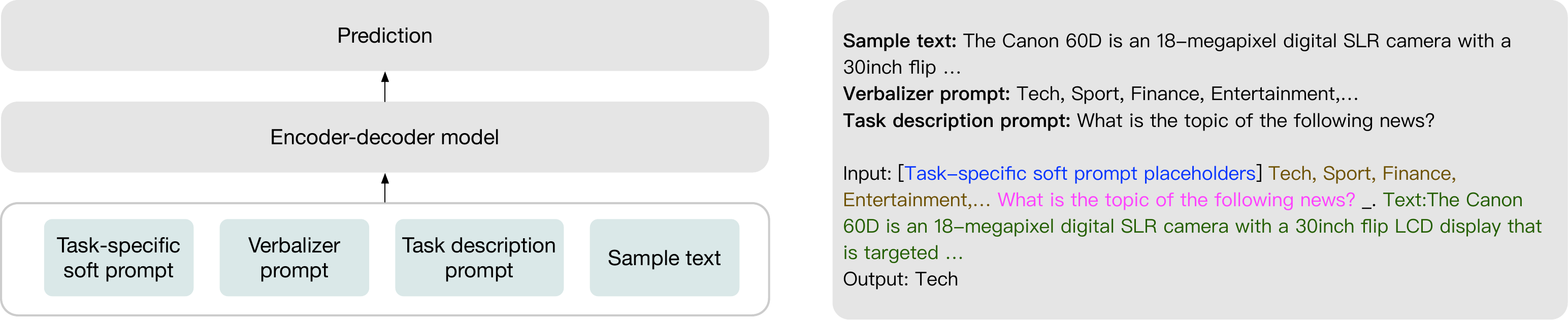}
    \caption{The hybrid prompt composed of task-specific soft prompt, verbalizer prompt and task description prompt.}
    \label{fig:prompt_design}
\end{figure*}

In this subsection, we describe the prompt design of our choice and some other tested variants.

In the simplest form of a prompt template $T$, the prompting method constructs $T$ by a handcrafted prompt $P$ and the text input sequence $X$: 
% \begin{equation}
$
    T = \{P, X, \textrm{[MASK]}\}
% \end{equation}
$
where $\textrm{[MASK]}$ is the blank that an answer should be filled in to complete the sentence. This is known as sentence in-filling.

As illustrated in Figure~\ref{fig:prompt_design}, our optimized prompt $P$ is further decomposed into three parts, $\mathcal{E}$, $\mathcal{V}$, and $\mathcal{D}$, where we have the task-specific soft prompt $\mathcal{E}$, the verbalizer prompt $\mathcal{V}$ and the task description prompt $\mathcal{D}$. As a result, our prompt template $T$ could be expressed as: 
\begin{equation}
    T = \{\mathcal{E}, \mathcal{V}, \mathcal{D}, X, \textrm{[MASK]}\}
\label{equ:prompt_design}
\end{equation}
To disentangle the task-specific and task-agnostic knowledge in multitask pretraining, we install a continuous prompt embedding as a prefix, which is referred as the task-specific soft prompt shown in Figure~\ref{fig:prompt_design}.

\begin{table}[t]
\footnotesize
    \centering
    \begin{tabular}{m{0.45in}<{\centering}m{0.55in}<{\centering}m{0.55in}<{\centering}m{0.55in}<{\centering}}
    \toprule
        & All & Seen & Unseen \\
    \midrule
    proposed          & 46.16($\uparrow$3.89) & 46.82($\uparrow$2.83) & 41.57($\uparrow$11.4) \\
    - $\mathcal{V}$    & 42.88($\uparrow$0.61) & 43.87($\downarrow$0.12) & 35.92($\uparrow$5.75) \\
    - $\mathcal{E}$    & 45.06($\uparrow$2.79) & 46.40($\uparrow$2.41) & 35.66($\uparrow$5.49) \\
    - $\mathcal{E, V}$ & 42.27 & 43.99 & 30.17 \\
    \bottomrule
    \end{tabular}
    \caption{Ablation results on the optimized prompt design. -$\mathcal{V}$: without the verbalizer prompt; - $\mathcal{E}$: without the task-specific soft prompt; - $\mathcal{E, V}$: without the verbalizer prompt and the task-specific soft prompt.}
    \label{tab:ablation_prompt_design}
\end{table}

We first validate the importance of including the task-specific soft prompt and the verbalizer prompt in our choice of prompt design, and then compare different methods to build new task-specific prompt embeddings. 
Ablation results on the optimized prompt design are shown in Table~\ref{tab:ablation_prompt_design}. 
We can see that task-specific soft prompts and verbalizer prompts are useful when applied separately, and can obtain an even greater gain of 4 points when applied combined by our \proposed. 

% \subsection{Build Task-specific Soft Prompts for Unseen Tasks}
% \label{appendix:soft_prompt}
For unseen tasks, we need to build task-specific soft prompts without any labeled sample. Firstly, we tune a classifier on the mixture of training data to tell the belongings of given texts, and for new samples in the test task, the classifier can predict the similarities of this sample to training tasks. 
Formally, for pretrained task $i$, we regard its task-specific prompt embedding as $\mathcal{E}_i$, the classifier output of training task $i$'s probability as $prob_i$. 
In our experiments, we have tried three methods to build the test task prompt embedding $\mathcal{E}_{new}$, they are ~\emph{weighted, top1 and random}.

1) \emph{weighted}. For the \emph{weighted}, we set $\mathcal{E}_{new}$ as a weighted average of pretrained task prompt embedding according to the probability, as
\begin{equation}
    \centering
    \mathcal{E}_{new} = \sum_{i=1}^N prob_i\times\mathcal{E}_i
    \label{eq:tpe1}
\end{equation}
Note that we can do the weighted average on the sample level, as well as the task level.

2) \emph{top1}. For the \emph{top1}, we assign the most similar task prompt embedding to the new task, as  
\begin{equation}
    \centering
    \begin{aligned}
    &\mathcal{E}_{new} = \mathcal{E}_k\\
    \textrm{where} ~~&k=\arg \max_{i}(prob_i),\, i\in N
    \end{aligned}
    \label{eq:tpe2}
\end{equation}

3) \emph{random}. For the \emph{random}, we initialize the task prompt embedding $\mathcal{E}_{new}$ randomly.

\begin{table}[t]
\footnotesize
    \centering
    \begin{tabular}{lcccc}
    \toprule
         & \tabincell{c}{none} &
        % \tabincell{c}{weighted avg \\all samples}   & 
        % \tabincell{c}{weighted avg \\per sample}   & 
        \tabincell{c}{weighted avg}   & 
        % \tabincell{c}{top1 \\ per sample}  & 
        \tabincell{c}{top1}  &
        \tabincell{c}{random init} \\ \midrule
        % \tabincell{c}{\proposed \\()}  \\ \hline
        All & 44.83  & 46.01 & 46.06 & \textbf{46.16}\\ 
        Seen & 46.67 & 46.77 & 46.79 & \textbf{46.82}\\ 
        Unseen & 31.98 & 40.65 & 40.95 & \textbf{41.57}\\ 
    \bottomrule
    \end{tabular}
    \caption{Ablation results on building new task-specific soft prompt embeddings.}
    \label{tab:ablation_task_prompt_embedding}
\end{table}

Ablation results are given in Table~\ref{tab:ablation_task_prompt_embedding}. Note that for \textit{weighted avg} and \textit{top1} we only report results of per sample, results with all samples are given in Table~\ref{tab:Detail_Table_for_Ablation_Studies_on_building_new_task_prompt_embedding}.
We can see that the winning approach is surprisingly \textit{random init}, and the direct uses of similar task prompt embeddings seen in training in various ways are slightly worse than \textit{random init}, and the worst performing method is \textit{none} as expected. 
To comprehend the results on \textit{random init} and \textit{top1}, we suppose that different tasks, though with similar input data distributions, still have different mappings $\mathcal{X}{\to}y$.
Therefore, it is often difficult to find the most proper task-specific soft prompt seen in the training phase for a new task in the zero-shot learning setting.

\subsection{Data Retrieval and Self-training}
\label{appendix:self-training}

To fully exploit unsupervised data, we take a self-training framework similar to~\cite{lee2013pseudo,du-etal-2021-self}. 
Given a supervised training set $D_{train}$ and an unlabeled dataset $D_{un}$, 
we will retrieve task-similar data from unsupervised corpus according to sentence embedding similarity, and the self-training process may repeat several times. 
For sentence embedding in retrieval, a pretrained BERT is finetuned on both unsupervised and supervised corpus using SimCSE~\cite{gao2021simcse}. 

The process of self-training is presented in Algorithm~\ref{alg:self-training-Framwork}, where $\mathcal{M}$ is the pretrained model, $T$ is the self-training epoch. For a specific task $i$, $D_{train}^i$ is the training set and $D_{un}^i$ is the unlabeled dataset. We note $D_{train}$ as the union of all training datasets and $D_{un}$ as the union of all unlabeled datasets.
\begin{table*}
\small
\centering
\begin{tabular}{lcccc}
\toprule
 & \multicolumn{2}{c}{Dev} & \multicolumn{2}{c}{Test} \\
Task & baseline & self-training & baseline & self-training \\ \midrule
NEWS AVG      & 86.49      & 87.45 ($\uparrow$0.96)    & 55.21      & 59.11 ($\uparrow$3.90)       \\ 
production AVG  & 81.84     & 81.94 ($\uparrow$0.10)  & 78.08     & 79.31 ($\uparrow$1.23) \\ 
\bottomrule
\end{tabular}
\caption{Experimental results on data retrieval + self-training}
\label{tab:Retrieval_self-training}
\end{table*}

\begin{algorithm}[t]
  \caption{Self-training}  
  \label{alg:self-training-Framwork}  
  \begin{algorithmic}[1]
    \Require  
      $\mathcal{M}$, $D_{un}$, $D_{train}$, $T$
    \Ensure  
      $\mathcal{M}^*$
    \State Init $D_{train}^* \gets D_{train}$
    \For{each $t\in [0,T]$} 
        \State $\mathcal{M}^*$ $\gets$ train $M$ on ${D^*}_{train}^i$
        \For{each task $i$}
        \State inference with $\mathcal{M}^*$ on $D_{un}^i$
        \State ${D^*}_{un}^i$ $\gets$ select samples in $D_{un}^i$ which are confident with $\mathcal{M}^*$ and make pseudo label,
        \State $D_{train}^* \gets D_{train}^* \cup {D^*}_{un}^i$ ,
        \EndFor
    \EndFor \\
    \Return $\mathcal{M}^*$;  
  \end{algorithmic}  
\end{algorithm}

% For news classification, we take 4 train datasets, 5 test datasets and 7 million unsupervsied sentences. For pruduction data, we take 122 train datasets, 130 test datasets and 20 million unsupervised sentences. Unsupervised sentences are randomly sampled from the pretraining corpus.
% Then we perform sentence encoding and retrieve task-similar data from the unsupervised sentences to build $D_{un}$. For sentence encoder, we finetuned a pretrained BERT with SimCSE on the mixture of unsupervised corpus and supervised data.
% For data retrieval, we use label average embedding as the query vector in the news task, and take the top 1K most similar sentences for each label. For production task, we use sentence embedding as the query vector and take the top 10 most similar sentences for each sentence.

We select new classification and production datasets to study the impact of data retrieval and self-training, considering similar data available in the unsupervised pretraining corpus. Results are summarized in Table~\ref{tab:Retrieval_self-training}. 
Self-training improves the validation set performance of 0.96 and 0.10 for NEWS and production tasks respectively, and improves the test zero-shot performance of 3.90 and 1.23. Self-training shows larger improvement on unseen tasks than training tasks. We explain that pseudo labeled data may increase the diversity of training data, resulting better zero-shot generalization abilities.

\begin{figure}[t]
    \centering
    \includegraphics[width=0.45\textwidth]{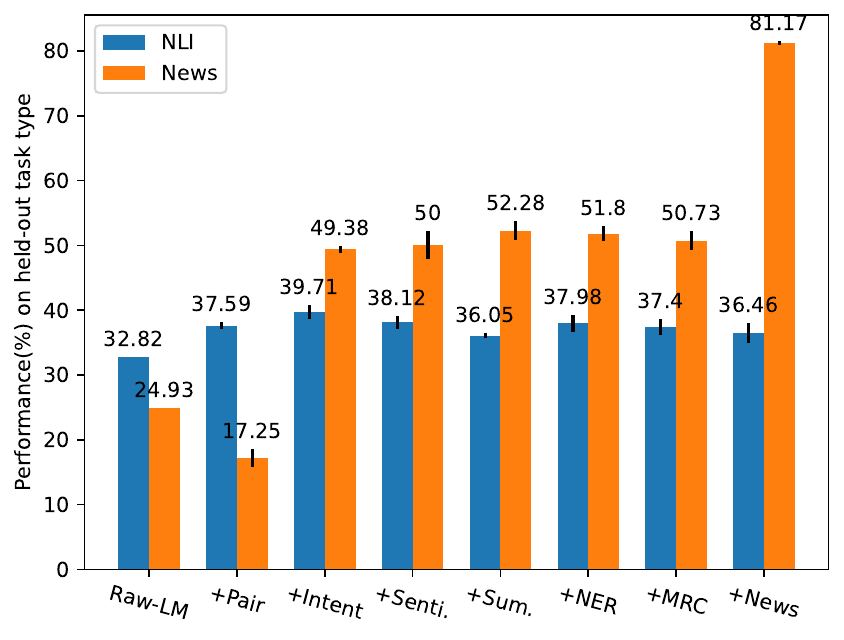}
    \caption{Zero-shot performance on NLI and NEWS with different held-out task types.}
    \label{fig:cross_task_type}
\end{figure}

\subsection{Effect of Cross Task Type Transfer}
\label{appendix:cross_type_transfer}
Following the previous works~\cite{wei2021finetuned,sanh2021multitask}, we study whether held-out task types can benefit from multitask prompted pretraining. 
Specifically, we choose NLI and NEWS as testing task types while other various datasets as training task types. 
We add different training tasks in sequence as shown in Figure~\ref{fig:cross_task_type}. 
For NEWS, the zero-shot performance increases from 17 to 49 by adding INTENT, while adding sentence pair (STS, QAM, PARA) tasks leads to a performance drop in 7 points. 
Other training task types such as SENTI, SUMM, NER and MRC only have marginal impacts on the performance. 
For sanity check, we add NEWS in the training phase at last and the performance increases from 50 to 81 as expected. 
The zero-shot performance on NLI rises from 32 to 37 by adding more sentence pair tasks, and then to 39 with INTENT, but other training tasks do not further boost the performance. 
In conclusion, we find that the zero-shot performance on held-out task types can only benefit from some task types, and more labeled data in other task clusters do not always guarantee continuous improvement.

In comparison, our main results on task scaling indicate that performance is boosted when the number of training tasks increases according to the fixed task distribution. Note that task distribution is orthogonal to scaling the task number. How to further improve zero-shot generalization by optimizing task distribution is left to future work.

% \balance

% \cleardoublepage

\begin{table*}[t]
\tiny
    \centering
    \begin{tabular}{lccccc}
    \toprule
        Task Type & Task & Train & Test & Metric \\ \midrule
        % \tabincell{c}{\proposed \\()}  \\ \hline
        \multirow{17}{*}{Sentiment Analysis (\textbf{SENTI})} & yf\_amazon & \checkmark & & Micro-F1 \\
        ~ & JD\_full & \checkmark & & Micro-F1  \\
        ~ & JD\_binary & \checkmark & & Micro-F1 \\
        ~ & waimai\_10k & \checkmark & & Micro-F1 \\
        ~ & online\_shopping\_10cats & & \checkmark & AUC \\
        ~ & ChnSentiCorp\_htl\_all & & \checkmark & AUC  \\
        ~ & nlpcc2014\_task2 & & \checkmark & AUC \\
        ~ & weibo\_senti\_100k & & \checkmark & AUC \\
        ~ & yf\_dianping & & \checkmark & Micro-F1 \\
        ~ & car\_sentiment & & \checkmark & Micro-F1 \\
        ~ & dmsc & & \checkmark & Micro-F1 \\
        ~ & simplifyweibo\_4 & & \checkmark & Micro-F1 \\
        ~ & NLPCC2014\_Weibo\_Emotion\_classification & & \checkmark & Micro-F1 \\
        ~ & nCoV\_100k & & \checkmark & Micro-F1 \\
        ~ & Internet\_News & & \checkmark & Micro-F1 \\
        ~ & BDCI2019 &  & \checkmark & Micro-F1 \\
        ~ & SMP2019\_ECISA &  & \checkmark & Micro-F1 \\ \midrule
        \multirow{9}{*}{News Classification(\textbf{NEWS})} & NLPCC2014\_LSHT\_sample & \checkmark & & Micro-F1 \\
        ~ & Chinanews & \checkmark &  & Micro-F1 \\
        ~ & CNSS & \checkmark & & Micro-F1 \\
        ~ & CNSE & \checkmark & & Micro-F1 \\
        ~ & THUCNews &  & \checkmark & Micro-F1 \\
        ~ & CCFBDCI2020 & & \checkmark & Micro-F1 \\
        ~ & tnews\_public & & \checkmark & Micro-F1 \\
        ~ & Ifeng &  & \checkmark & Micro-F1 \\
        ~ & nlpcc2017\_news\_headline\_categorization &  & \checkmark & Micro-F1 \\ \midrule
        \multirow{4}{*}{Intent Classification (\textbf{INTENT})} & nlpcc2018\_slu & \checkmark & & Micro-F1 \\
        ~ & catslu\_traindev & & \checkmark & Micro-F1 \\
        ~ & e2e\_dials &  & \checkmark & Micro-F1 \\
        ~ & intent\_classification &  & \checkmark & Micro-F1 \\ \midrule
        \multirow{2}{*}{Natural language inference (\textbf{NLI})} & cmnli\_public & \checkmark & & Micro-F1 \\
        ~ & ocnli\_public &  & \checkmark & Micro-F1 \\ \midrule
        \multirow{13}{*}{Sentence Similarity (\textbf{STS})} & LCQMC & \checkmark & & AUC \\
        ~ & bq\_corpus & \checkmark & & AUC \\
        ~ & sohu\_sts\_A\_sl & \checkmark & & AUC \\
        ~ & afqmc\_public & & \checkmark & AUC \\
        ~ & phoenix\_pair & & \checkmark & AUC \\
        ~ & sohu-sts-A-ll & & \checkmark & AUC \\
        ~ & sohu-sts-A-ss & & \checkmark & AUC \\
        ~ & sohu-sts-B-ll & & \checkmark & AUC \\
        ~ & sohu-sts-B-sl & & \checkmark & AUC \\
        ~ & sohu-sts-B-ss & & \checkmark & AUC \\
        ~ & CBLUE-CHIP-STS & & \checkmark & AUC \\
        ~ & CBLUE-KUAKE-QTR & & \checkmark & Micro-F1 \\
        ~ & CBLUE-KUAKE-QQR & & \checkmark & Micro-F1 \\ \midrule
        \multirow{1}{*}{Paraphrase (\textbf{PARA})} & PAWS-X & & \checkmark & AUC \\ \midrule
        \multirow{1}{*}{Question Answer Matching (\textbf{QAM})} & nlpcc2016-dbqa & & \checkmark & AUC \\ \midrule
        \multirow{10}{*}{\makecell[l]{Machine Reading Comprehension \\ Question Answering (\textbf{MRC})}} & c3\_public &\checkmark &  & F1 \\
        ~ & DuReader\_robust & \checkmark &  & F1 \\
        ~ & DuReader\_checklist & \checkmark &  & F1 \\
        ~ & DuReader\_yesno & \checkmark &  & F1 \\
        ~ & dureader & \checkmark &  & F1 \\
        ~ & cmrc2018\_public & & \checkmark & F1 \\
        ~ & DRCD & & \checkmark & F1 \\
        ~ & CCF2020-BDCI-QA & & \checkmark & F1 \\
        ~ & CAIL2019-QA & & \checkmark & F1 \\
        ~ & CAIL2020-QA & & \checkmark & F1 \\ \midrule
        \multirow{9}{*}{Name Entity Recognition (\textbf{NER})} & BosonNLP\_NER\_6C & \checkmark & & Pos-F1 \\
        ~ & cluener\_public & \checkmark & & Pos-F1 \\
        ~ & RENMIN\_NER & \checkmark & & Pos-F1 \\
        ~ & msra\_ner & & \checkmark & Pos-F1 \\
        ~ & weibo\_ner & & \checkmark & Pos-F1 \\
        ~ & nlpcc2020-AutoIE & & \checkmark & Pos-F1 \\
        ~ & CCF2020-BDCI-NER & & \checkmark & Pos-F1 \\
        ~ & CMeEE & & \checkmark & Pos-F1 \\
        ~ & SanWen-ner & & \checkmark & Pos-F1 \\ \midrule
        \multirow{9}{*}{Summarization (\textbf{SUMM})} & LCSTS & \checkmark & & ROUGE \\
        ~ & NLPCC2017 & \checkmark & & ROUGE \\
        ~ & SHENCE & \checkmark & & ROUGE \\
        ~ & NLPCC2015 & & \checkmark & ROUGE \\
        ~ & CAIL2020 & & \checkmark & ROUGE \\
        ~ & WANFANG & & \checkmark & ROUGE \\
        ~ & CSL\_SUMM & & \checkmark & ROUGE \\
        ~ & EDU\_SUMM & & \checkmark & ROUGE \\
        ~ & WEIBO & & \checkmark & ROUGE \\ \midrule
        \multirow{3}{*}{Keywords (\textbf{KEYS})} & COTE-BD & & \checkmark & F1 \\
        ~ & COTE-MFW &  & \checkmark & F1 \\
        ~ & COTE-DP &  & \checkmark & F1 \\ \midrule
        \multirow{1}{*}{Winograd Schema Challenge (\textbf{WSC})} & cluewsc2020\_public &  & \checkmark & AUC \\ \midrule
        \multirow{1}{*}{App Classification (\textbf{APP})} & iflytek\_public &  & \checkmark & Micro-F1 \\ \midrule
        \multirow{2}{*}{Production Datasets} & 800 datasets for training & \checkmark & & AUC \\
        ~ & 230 datasets for testing &  & \checkmark & AUC \\
        \bottomrule
        
    \end{tabular}
    \caption{Summary of collected datasets}
    \label{tab:summary_datasets}
\end{table*}

\subsection{Hard Prompt Examples}
In this section, we provide details of hard prompts used in this paper.
For tasks within each Chinese task cluster, we use similar handcrafted prompts as shown in
% Table~\ref{tab:production}\ref{tab:appendix_SENTI}\ref{tab:appendix_News}\ref{tab:appendix_News}\ref{tab:appendix_MRC}\ref{tab:appendix_NLI}\ref{tab:appendix_STS}\ref{tab:appendix_WSC}\ref{tab:appendix_NER}\ref{tab:appendix_SUM} . 
Table~\ref{tab:production} $\sim$ \ref{tab:appendix_SUM} .
We use both \textit{prefix prompts} and \textit{cloze prompts}. For text classification clusters such as SENTI, NEWS, [X] denotes the sample text. For sentence pair task clusters such as NLI, STS, [X1] denotes the first sample sentence and [X2] is the second sample sentence. For cluster MRC, [X1] denotes the coupus and [X2] denotes the question. For cluster SUM, [X] denotes the coupus, and a similar prompt form is applied for KEYS. For NER, [X1] is the sample text and [X2] denotes the target entity type. For WSC, [X1] is the sample text and [X2] is the pronoun. For all prompts mentioned above, '\_' denotes the target position to fill in the answer.

% For each English datasets, we conduct our Genetic Prompt Search method with prompts initialized from T0 paper \citep{sanh2021multitask}.
% In Table \ref{tab:en-gps}, we gives the generated prompts that achieve best performance on each datasets.

\begin{table*}[!h]
	\small
    \centering
    \begin{tabular}{p{2cm}p{10.5cm}p{2.5cm}}
        \toprule
         Task Type & Prompts & label \\ \midrule
         \multirow{4}{*}{\textbf{Objection}} & \begin{CJK}{UTF8}{gbsn}Prompt: 这句话：[X]。上文是否体现了客户对公司不信任？回答： \end{CJK} & \multirow{4}{*}{\begin{CJK}{UTF8}{gbsn}是(Yes)/不是(No)\end{CJK}} \\
         ~ & \begin{CJK}{UTF8}{gbsn}X: 你们是什么公司啊？我从来没听说过你们。\end{CJK} & ~ \\
         ~ & Prompt: This sentence: [X]. Does the customer show objection about the company? Answer: & ~ \\
         ~ & X: What kind of company are yours? I have never heard of it. & ~ \\
         \midrule
         \multirow{4}{*}{\textbf{Profile}} & \begin{CJK}{UTF8}{gbsn}Prompt: 这句话：[X]。客户是在询问用药后的效果吗？回答： \end{CJK} & \multirow{4}{*}{\begin{CJK}{UTF8}{gbsn}是(Yes)/不是(No)\end{CJK}} \\
         ~ & \begin{CJK}{UTF8}{gbsn}X: 吃了以后的主要作用是什么？。\end{CJK} & ~ \\
         ~ & Prompt: This sentence: [X]. Is the customer asking about the influences of taking the medicine? Answer: & ~ \\
         ~ & X: What is the main effect after taking this? & ~ \\
         \midrule
         \multirow{6}{*}{\makecell[l]{\textbf{Acception}}} & \begin{CJK}{UTF8}{gbsn}Prompt: 关于电子保单查看， “[X1]”上文销售采纳了与系统推荐“[X2]”相似的描述吗？回答： \end{CJK} & ~ \\
         ~ & \begin{CJK}{UTF8}{gbsn}X1: 让我看一下啊这个您电子版保单这块咱们有接收到吗？ \end{CJK} & ~ \\
         ~ & \begin{CJK}{UTF8}{gbsn}X2: 您的这个电子保单合同有没有收到呢？\end{CJK} & \begin{CJK}{UTF8}{gbsn}采纳(Accept)/\end{CJK} \\
         ~ & Prompt: About electronic insurance policy, Does the salesman say "[X1]" accept the system given expression "[X2]"? Answer: & \begin{CJK}{UTF8}{gbsn}没有(No)\end{CJK} \\
         ~ & X1: Let me see. Did you received our electronic version of insurance policy?  & ~ \\
         ~ & X2: Have you received this electronic policy contract? & ~ \\
         \midrule
         \multirow{4}{*}{\textbf{Violation}} & \begin{CJK}{UTF8}{gbsn}Prompt: 这句话：[X]。上文是否体现了坐席私自承诺客户可以随时退款？回答： \end{CJK} & \multirow{4}{*}{\begin{CJK}{UTF8}{gbsn}是(Yes)/不是(No)\end{CJK}} \\
         ~ & \begin{CJK}{UTF8}{gbsn}X: 如果说觉得感觉不太满意的话，你可以直接申请退款。一个月之内，申请退款。\end{CJK} & ~ \\
         ~ & Prompt: This sentence: [X]. Does the customer service privately promise that the customer can refund at any time? Answer: & ~ \\
         ~ & X: If you feel unsatisfied, you can directly apply for a refund. Within one month, apply for a refund. & ~ \\
         \midrule
         \multirow{6}{*}{\makecell[l]{\textbf{Mention}}} & \begin{CJK}{UTF8}{gbsn}Prompt: 关于保单理赔， “[X1]”是销售提及的内容与文本“[X2]”相似吗？回答： \end{CJK} & ~ \\
         ~ & \begin{CJK}{UTF8}{gbsn}X1: 55种轻症疾病和保险公司达成理赔协议之后7到100个工作日，一次性就把这个钱赔给你了。 \end{CJK} & ~ \\
         ~ & \begin{CJK}{UTF8}{gbsn}X2: 二级及以上公立医院医生的诊断报告啊就可以申请理赔。保险公司呢是直接一次性给到我们100万块钱去看病了。\end{CJK} & \begin{CJK}{UTF8}{gbsn}相似(similar)/\end{CJK} \\
         ~ & Prompt: About insurance claim, Does the salesman say "[X1]" mentioned a similar description as "[X2]"? Answer: & \begin{CJK}{UTF8}{gbsn}不同(different)\end{CJK} \\
         ~ & X1: For 55 mild disease, it will cost 7 to 100 working days after reaching a claim settlement agreement with the insurance company, after that, the money will be paid to you. & ~ \\
         ~ & X2: You can apply for a claim with the diagnosis report of a doctor in a public hospital of level 2 or above. The insurance company will gave you 1 million yuan directly for the disease. & ~ \\
         \midrule
         \multirow{4}{*}{\textbf{Execution}} & \begin{CJK}{UTF8}{gbsn}Prompt:  这句话：[X]。上文坐席是否告知客户存在优惠价格？回答： \end{CJK} & \multirow{4}{*}{\begin{CJK}{UTF8}{gbsn}是(Yes)/不是(No)\end{CJK}} \\
         ~ & \begin{CJK}{UTF8}{gbsn}X: 咱们现在也是有优惠活动的，为何不趁着优惠活动把身体调整一下呢？\end{CJK} & ~ \\
         ~ & Prompt: This sentence: [X]. Does the salesman told customer there are discount price? Answer: & ~ \\
         ~ & X: We have a discount price right now, why not take a change with this discounts? & ~ \\
        %  \midrule
        %  \multirow{6}{*}{\textbf{Execution}} & \begin{CJK}{UTF8}{gbsn}Prompt: 这句话 [X]。上文坐席是否告知客户存在优惠价格？回答： \end{CJK} & ~ \\
        %  ~ & \begin{CJK}{UTF8}{gbsn}X: 咱们现在也是有优惠活动的，为何不趁着优惠活动把身体调整一下呢。。 \end{CJK}& \begin{CJK}{UTF8}{gbsn}相似(similar)/\end{CJK} \\
        %  ~ & Prompt: Do "[X1]" and "[X2]" mean the same thing? Answer: & \begin{CJK}{UTF8}{gbsn}不同(different)\end{CJK} \\
        %  ~ & X1: If you sign up today, you can participate in the activity that pay 200 yuan and get extra 200 yuan reward. & ~ \\
        %  ~ & X2: I have received. I don't agree. My account balance is insufficient. & ~ \\
         \bottomrule
         
    \end{tabular}
    \caption{Illustrations of examples in our production datasets.}
    \label{tab:production}
\end{table*}

\begin{table*}
	\small
    \centering
    \begin{tabular}{|p{15cm}|}
        % \toprule
        % \hline 
        % \textbf{car\_sentiment}\\
        \hline 
        \textbf{Handcrafted} \\
         \begin{CJK}{UTF8}{gbsn}Prompt: “[X]”这句汽车评论的态度是什么？\_。\end{CJK} \\  
            Prompt: "[X]", What is the attitude of this car review ?\_ \\ 
            \begin{CJK}{UTF8}{gbsn}X: 动力还可以因为搭载cvt变速箱起步发动机转速比较好。\end{CJK} \\
            X: Power can also be equipped with a CVT gearbox to start the engine speed is better. \\
        \textbf{Augmentation} \\
         \begin{CJK}{UTF8}{gbsn}Prompt: “[X]”如果这个评论的作者是客观的,那么请问,这个评论的内容是什么回答：？\_。\end{CJK} \\  
            Prompt: "[X]", If the author of this comment is objective, what is the content of this comment reply: \_ \\ 
        \textbf{Verbalizer} \\ 
         \begin{CJK}{UTF8}{gbsn}积极(Positive)/消极(Negative)\end{CJK}\\
         \hline 
         \textbf{Target} \\
          \begin{CJK}{UTF8}{gbsn}积极(Positive)\end{CJK}\\
         \hline
    \end{tabular}
    \caption{Illustrations of prompts in Sentiment Analysis.}
    \label{tab:appendix_SENTI}
\end{table*}

\begin{table*}
	\small
    \centering
    \begin{tabular}{|p{15cm}|}
        % \toprule
        % \hline 
        % \textbf{car\_sentiment}\\
        \hline 
        \textbf{Handcrafted} \\
         \begin{CJK}{UTF8}{gbsn}Prompt: 以下这篇新闻是关于什么主题的？\_。新闻：[X]\end{CJK} \\  
            Prompt: What is the topic of the following news?\_. News text: [X] \\ 
            \begin{CJK}{UTF8}{gbsn}X: 1800万像素单反 佳能60D套机降至9700元   作者：陈 【北京行情】 佳能60D(资料 报价 图片 论坛)是一款拥有1800万像素成像能力，搭载3英寸可翻转LCD显示屏，定位于中端市场的数码单反相机。... 作为佳能畅销单反50D的继承者，佳能EOS 60D对于想拥有一台中端单反相机的用户无疑是一个不错的选择。\end{CJK} \\
            X: The Canon 60D is an 18-megapixel digital SLR camera with a 3-inch flip LCD display that is targeted at the mid-market. ... The successor to Canon's best-selling DSLR 50D, the Canon EOS 60D is a good choice for anyone who wants a mid-range DSLR camera. \\
        \textbf{Augmentation} \\
         \begin{CJK}{UTF8}{gbsn}Prompt: ‘新闻文本’是谁写的?回答：\_。“[X]”\end{CJK} \\  
            Prompt: Who wrote the 'news text'? Answer: \_. "[X]" \\ 
        \textbf{Verbalizer} \\ 
         \begin{CJK}{UTF8}{gbsn}科技(Technology)/体育(Sport)/财经(Finance)/娱乐(Entertainment)/..\end{CJK}\\
         \hline 
         \textbf{Target} \\
          \begin{CJK}{UTF8}{gbsn}科技(Technology)\end{CJK}\\
         \hline
    \end{tabular}
    \caption{Illustrations of prompts in News Classification.}
    \label{tab:appendix_News}
\end{table*}

\begin{table*}
	\small
    \centering
    \begin{tabular}{|p{15cm}|}
        % \toprule
        % \hline 
        % \textbf{car\_sentiment}\\
        \hline 
        \textbf{Handcrafted} \\
         \begin{CJK}{UTF8}{gbsn}Prompt: 文章：[X1] 问题：[X2] 回答：\_。\end{CJK} \\  
            Prompt: Corpus: [X1] Question: [X2] Answer:\_. \\ 
            \begin{CJK}{UTF8}{gbsn}X1: 微信一天最多能转多少钱:这个没有限制吧，到账时间长。纠正下其他网友的回答，微信转账是有限额的。用微信零钱转账最高可以1W元，用银行卡支付就要以银行的额度为准了，最高可以转账5W元。请采纳哦。\end{CJK} \\
            \begin{CJK}{UTF8}{gbsn}X2: 微信一天最多能转多少钱？ \end{CJK}\\
            X1: Micro letter a day how much money can transfer: there is no limit to it, long to the account. To correct other netizens' answers, wechat transfers are limited. The maximum amount can be 1W yuan with wechat change, and the maximum amount can be 5W yuan for bank card payment. Please adopt it. \\
            X2: How much money can wechat transfer at most a day? \\ 
        \textbf{Augmentation} \\
         \begin{CJK}{UTF8}{gbsn}Prompt: 他们是怎么猜出来的?文章：[X1] 问题：[X2] 回答：\_。\end{CJK} \\  
            Prompt: How did they figure that out? Corpus: [X1] Question: [X2] answer: \_ \\ 
         \hline 
         \textbf{Target} \\
          \begin{CJK}{UTF8}{gbsn}微信转账是有限额的。用微信零钱转账最高可以1W元，用银行卡支付就要以银行的额度为准了，最高可以转账5W元\end{CJK}\\
          Wechat transfers are limited. The maximum amount can be 1W yuan with wechat change, and the maximum amount can be 5W yuan for bank card payment. \\
         \hline
    \end{tabular}
    \caption{Illustrations of prompts in Machine Reading Comprehension Question Answering.}
    \label{tab:appendix_MRC}
\end{table*}

\begin{table*}
	\small
    \centering
    \begin{tabular}{|p{15cm}|}
        % \toprule
        % \hline 
        % \textbf{car\_sentiment}\\
        \hline 
        \textbf{Handcrafted} \\
         \begin{CJK}{UTF8}{gbsn}Prompt: 在通用领域中，第一句话：“[X1]”第二句话：“[X2]”的逻辑关系是什么？回答：\_。\end{CJK} \\  
            Prompt: In the general context, What is the logical relationship between the first sentence "[X1]" and the second sentence "[X2]". Answer: \_.  \\ 
            \begin{CJK}{UTF8}{gbsn}X1: 等他回来,我们就出去吃啊。\end{CJK} \\
            X1: When he gets back, we'll eat out. \\
            \begin{CJK}{UTF8}{gbsn}X2: 我们在等他。\end{CJK} \\
            X2: We are waiting for him. \\
        \textbf{Augmentation} \\
         \begin{CJK}{UTF8}{gbsn}Prompt: 这两句话是如何组合在一起的?回答：\_。第一句话：“[X1]”，第二句话：“[X2]”\end{CJK} \\  
            Prompt: How do these two sentences go together? Answer: \_. the first sentence: "[X1]", the second sentence: "[X2]".\\ 
        \textbf{Verbalizer} \\ 
         \begin{CJK}{UTF8}{gbsn}相反(contradiction)/中性(neutral)/一致(entailment)\end{CJK}\\
         \hline 
         \textbf{Target} \\
          \begin{CJK}{UTF8}{gbsn}一致(entailment)\end{CJK}\\
         \hline
    \end{tabular}
    \caption{Illustrations of prompts in Natural Language Inference.}
    \label{tab:appendix_NLI}
\end{table*}

\begin{table*}
	\small
    \centering
    \begin{tabular}{|p{15cm}|}
        % \toprule
        % \hline 
        % \textbf{car\_sentiment}\\
        \hline 
        \textbf{Handcrafted} \\
         \begin{CJK}{UTF8}{gbsn}Prompt: 在金融领域中，第一句话：“[X1]”第二句话：“[X2]”这两句话含义 \_。\end{CJK} \\  
            Prompt: In finance context, the first sentence: "[X1]" the second sentence: "[X2]", the meaning of these two sentences is \_.  \\ 
            \begin{CJK}{UTF8}{gbsn}X1: 花呗支持高铁票支付吗？\end{CJK} \\
            X1: Does Huabei support high-speed rail ticket payment? \\
            \begin{CJK}{UTF8}{gbsn}X2: 为什么不支持花呗付款？\end{CJK} \\
            X2: Why not support the payment of Huabei? \\
        \textbf{Augmentation} \\
         \begin{CJK}{UTF8}{gbsn}Prompt: 它们之间的关系是怎样的?回答：\_。第一句话：“[X1]”，第二句话：“[X2]”\end{CJK} \\  
            Prompt: What is the relationship between them? Answer: \_. the first sentence: "[X1]", the second sentence: "[X2]".\\ 
        \textbf{Verbalizer} \\ 
         \begin{CJK}{UTF8}{gbsn}相似(similar)/不同(different)\end{CJK}\\
         \hline 
         \textbf{Target} \\
          \begin{CJK}{UTF8}{gbsn}不同(different)\end{CJK}\\
         \hline
    \end{tabular}
    \caption{Illustrations of prompts in Sentence Similarity.}
    \label{tab:appendix_STS}
\end{table*}

\begin{table*}
	\small
    \centering
    \begin{tabular}{|p{15cm}|}
        % \toprule
        % \hline 
        % \textbf{car\_sentiment}\\
        \hline 
        \textbf{Handcrafted} \\
         \begin{CJK}{UTF8}{gbsn}Prompt: 对于句子：[X1] 代词：[X2] 指代的是：[X3] 吗？回答：\_。\end{CJK} \\  
            Prompt: In the sentence: [X1], does the pronoun [X2] refer to [X3]? Answer: \_. \\ 
            \begin{CJK}{UTF8}{gbsn}X1: 满银的老祖上曾经当过“拔贡”。先人手里在这一带有过些名望。到他祖父这代就把一点家业败光了。\end{CJK} \\
            \begin{CJK}{UTF8}{gbsn}X2: 他 \end{CJK}\\
            \begin{CJK}{UTF8}{gbsn}X3: 满银 \end{CJK}\\
            X1: The old ancestor of Manyin used to be "baogong". There was some renown in the hands of our ancestors. By his grandfather's generation the family business had been wiped out. \\
            X2: he \\ 
            X3: Manyin \\ 
        \textbf{Augmentation} \\
         \begin{CJK}{UTF8}{gbsn}Prompt: 第二句话中,有两个“它"：[X1] 其中：[X2]指的\_[X3]。\end{CJK} \\  
            Prompt: In the second sentence, there are two "it" s: [X1] among this sentence: [X2] refer to [X3]? \_ \\ 
        \textbf{Verbalizer} \\ 
         \begin{CJK}{UTF8}{gbsn}是(yes)/不是(no)\end{CJK}\\
         \hline 
         \textbf{Target} \\
          \begin{CJK}{UTF8}{gbsn}是(yes)\end{CJK}\\
         \hline
    \end{tabular}
    \caption{Illustrations of prompts in Winograd Schema Chanllenge.}
    \label{tab:appendix_WSC}
\end{table*}

\begin{table*}
	\small
    \centering
    \begin{tabular}{|p{15cm}|}
        % \toprule
        % \hline 
        % \textbf{car\_sentiment}\\
        \hline 
        \textbf{Handcrafted} \\
         \begin{CJK}{UTF8}{gbsn}Prompt: 报纸文本：[X1]中有哪些属于[X2]？回答\end{CJK} \\  
            Prompt: Text from newspaper : Which words of [X1] belong to [X2]? Answer: \_. \\ 
            \begin{CJK}{UTF8}{gbsn}X1: 相比之下，青岛海牛队和广州松日队的雨中之战虽然也是0∶0，但乏善可陈。\end{CJK} \\
            \begin{CJK}{UTF8}{gbsn}X2: 机构名 \end{CJK}\\
            X1: In contrast, although the raining war between Qingdao manatee team and Guangzhou songri team is also 0:0, but it is too lackluster. \\
            X2: organization \\ 
        \textbf{Augmentation} \\
         \begin{CJK}{UTF8}{gbsn}Prompt: 回答：\_。文本[X1] 报纸文本中的[X2]类别的实体是由哪些部分构成的？\end{CJK} \\  
            Prompt: Answer: \_. Text from newspaper: [X1]. Which parts make up the entities of the [X2] category in newspaper text?\\
         \hline 
         \textbf{Target} \\
          \begin{CJK}{UTF8}{gbsn} 青岛海牛队，广州松日队 \end{CJK}\\
          Qingdao manatee team, Guangzhou songri team \\
         \hline
    \end{tabular}
    \caption{Illustrations of prompts in Name Entity Recognition. Each example is extended to N instances, where N is the number of possible entity type. For each entity type, we ask the model to predict corresponding entities presented in the given text. The ground truth is "blank" if there is no entity of that type in the sentence.}
% {\begin{CJK}{UTF8}{gbsn}\shortstack{没有} \end{CJK}}
    \label{tab:appendix_NER}
\end{table*}

\begin{table*}
	\small
    \centering
    \begin{tabular}{|p{15cm}|}
        % \toprule
        % \hline 
        % \textbf{car\_sentiment}\\
        \hline 
        \textbf{Handcrafted} \\
         \begin{CJK}{UTF8}{gbsn}Prompt: [X]，这个教育相关的文本的摘要为：\_。\end{CJK} \\  
            Prompt: [X], A summary of this education-related text: \_. \\ 
            \begin{CJK}{UTF8}{gbsn}X: 中新网2月25日电 据外媒报道，意大利一名小女孩嘉比是一位动物爱好者，她经常拿自己的零食和家里的剩菜喂乌鸦，因此而收到了乌鸦送的“礼物”。据报道，嘉比经常用花生、狗粮和一些剩菜喂乌鸦，她表示，自己不是为了获得奖励而做这些，而是因为她喜欢自然。最近，乌鸦经常衔一些亮晶晶的东西给她，里面通常是些钮扣、文具和五金之类的小东西，有几次她还收到耳环，乌鸦甚至帮她妈妈把遗失的相机盖找了回去。禽鸟专家表示，乌鸦确实有和人类交朋友的能力，所以乌鸦报恩不是小女孩的想象。\end{CJK} \\
            X: China News on February 25: Gabi, an Italian girl who loves animals, has received a gift from a crow for feeding her snacks and family leftovers, foreign media reported. Gaby reportedly regularly feeds the crows peanuts, dog food and some leftovers, and she said she does not ask a reward but because she loves nature. Lately, they've been bringing her shiny things, usually buttons, stationery and hardware. In a few cases, she's received earrings. They even helped her mother find the cover of a camera she'd lost. According to bird experts, crows do have the ability to make friends with humans, so it's not a little girl's imagination for them to return the favor. \\
        \textbf{Augmentation} \\
         \begin{CJK}{UTF8}{gbsn}Prompt: [X] 这个领域的领域词典中收录的单词,应该是\_。\end{CJK} \\
            Prompt: [X] The words in the domain dictionary of this field should be \_. \\ 
         \hline 
         \textbf{Target} \\
          \begin{CJK}{UTF8}{gbsn}意大利女童用零食喂乌鸦，乌鸦送“礼物”报恩"\end{CJK}\\
          Talian girl feeds snacks to crows who return kindness with 'gifts' \\
         \hline
    \end{tabular}
    \caption{Illustrations of prompts in Summarization.}
    \label{tab:appendix_SUM}
\end{table*}

\subsection{Detailed Experimental Results}
Detailed ablation results of each testing task are presented in Table~\ref{tab:Detail_Table_for_Ablation_Studies}$\sim$\ref{tab:Detail_Table_for_Ablation_Studies_on_building_new_task_prompt_embedding}.

\begin{table*}[t]
\footnotesize
    \centering
    \begin{tabular}{lccccc}
    \toprule
        Model & - $\mathcal{E, V}$ & - $\mathcal{V}$ & - $\mathcal{E}$ & 
        \tabincell{c}{\proposed} & + MLM \\ \midrule
        % \tabincell{c}{\proposed \\()}  \\ \hline
        Total Scores* & $42.27_{(0.34)}$ & $42.88_{(0.55)}$ & $45.06_{(0.69)}$ & $46.16_{(0.54)}$ & $47.43_{(0.76)}$\\ \midrule
        online\_shopping\_10cats & $96.11_{(0.31)}$ & $96.06_{(0.27)}$ & $95.55_{(0.31)}$ & $95.72_{(0.27)}$ & $95.90_{(0.24)}$\\
        ChnSentiCorp\_htl\_all & $93.80_{(0.51)}$ & $93.75_{(0.57)}$ & $93.44_{(0.47)}$ & $93.45_{(0.38)}$ & $93.98_{(0.55)}$\\
        nlpcc2014\_task2 & $79.05_{(0.81)}$ & $80.42_{(0.49)}$ & $80.28_{(0.64)}$ & $80.12_{(0.24)}$ & $80.49_{(0.41)}$\\
        yf\_dianping & $37.27_{(2.66)}$ & $37.27_{(3.85)}$ & $45.11_{(5.41)}$ & $44.87_{(4.48)}$ & $43.89_{(2.51)}$\\
        car\_sentiment & $23.98_{(0.57)}$ & $30.49_{(5.57)}$ & $24.38_{(1.64)}$ & $25.80_{(3.41)}$ & $25.63_{(1.70)}$\\
        dmsc & $34.25_{(2.13)}$ & $36.94_{(2.65)}$ & $37.16_{(3.73)}$ & $37.88_{(2.31)}$ & $36.97_{(3.08)}$\\
        weibo\_senti\_100k & $86.48_{(0.58)}$ & $86.39_{(1.99)}$ & $84.23_{(1.00)}$ & $85.89_{(1.22)}$ & $86.48_{(1.55)}$\\
        simplifyweibo\_4 & $18.70_{(2.20)}$ & $20.38_{(2.23)}$ & $44.58_{(1.20)}$ & $38.87_{(2.06)}$ & $42.66_{(4.60)}$\\
        NLPCC2014\_Weibo\_Emotion\_classification & $37.57_{(1.39)}$ & $38.90_{(1.20)}$ & $40.56_{(0.93)}$ & $41.21_{(1.08)}$ & $41.28_{(1.69)}$\\
        nCoV\_100k & $34.11_{(0.53)}$ & $33.62_{(1.59)}$ & $33.20_{(2.00)}$ & $34.82_{(1.35)}$ & $34.91_{(0.49)}$\\
        Internet\_News & $53.61_{(2.23)}$ & $48.99_{(1.95)}$ & $52.42_{(10.39)}$ & $55.20_{(8.58)}$ & $56.92_{(2.78)}$\\
        BDCI2019 & $26.91_{(5.09)}$ & $22.53_{(3.45)}$ & $29.75_{(5.22)}$ & $36.53_{(5.45)}$ & $32.81_{(3.04)}$\\
        SMP2019\_ECISA & $38.18_{(1.25)}$ & $36.44_{(1.51)}$ & $35.71_{(2.76)}$ & $38.44_{(1.87)}$ & $38.46_{(0.33)}$\\
        THUCNews & $47.43_{(2.77)}$ & $51.45_{(3.98)}$ & $66.06_{(2.14)}$ & $65.86_{(2.93)}$ & $68.66_{(1.29)}$\\
        CCFBDCI2020 & $71.92_{(0.98)}$ & $69.54_{(3.55)}$ & $74.78_{(4.00)}$ & $75.93_{(4.21)}$ & $80.50_{(1.68)}$\\
        tnews\_public & $35.10_{(1.14)}$ & $34.23_{(3.66)}$ & $46.67_{(1.49)}$ & $46.35_{(1.50)}$ & $49.90_{(1.36)}$\\
        Ifeng & $60.41_{(1.97)}$ & $57.96_{(4.12)}$ & $61.32_{(0.94)}$ & $62.79_{(1.21)}$ & $63.04_{(2.27)}$\\
        nlpcc2017\_news\_headline\_categorization & $33.00_{(1.67)}$ & $33.52_{(2.52)}$ & $47.56_{(1.72)}$ & $47.14_{(1.37)}$ & $50.26_{(1.43)}$\\
        catslu\_traindev & $90.79_{(0.56)}$ & $91.59_{(0.80)}$ & $90.45_{(0.43)}$ & $91.33_{(0.54)}$ & $90.48_{(0.78)}$\\
        e2e\_dials & $69.20_{(2.92)}$ & $67.27_{(4.14)}$ & $82.02_{(2.02)}$ & $86.39_{(5.50)}$ & $88.44_{(5.28)}$\\
        intent\_classification & $20.41_{(1.05)}$ & $24.99_{(0.52)}$ & $28.47_{(1.47)}$ & $34.37_{(4.38)}$ & $33.64_{(3.84)}$\\
        ocnli\_public & $45.60_{(1.19)}$ & $47.60_{(0.16)}$ & $47.70_{(1.20)}$ & $47.16_{(2.09)}$ & $46.16_{(1.87)}$\\
        afqmc\_public & $63.40_{(0.79)}$ & $64.37_{(0.57)}$ & $63.63_{(0.91)}$ & $63.52_{(0.88)}$ & $64.60_{(0.49)}$\\
        phoenix\_pair & $98.90_{(0.22)}$ & $99.28_{(0.30)}$ & $98.77_{(0.44)}$ & $98.99_{(0.17)}$ & $98.99_{(0.24)}$\\
        sohu-sts-A-ll & $64.65_{(0.60)}$ & $64.04_{(0.97)}$ & $64.21_{(0.50)}$ & $65.44_{(0.72)}$ & $65.92_{(0.78)}$\\
        sohu-sts-A-ss & $70.91_{(0.37)}$ & $71.83_{(1.56)}$ & $69.88_{(1.34)}$ & $70.70_{(0.74)}$ & $70.80_{(0.46)}$\\
        sohu-sts-B-ll & $60.32_{(1.69)}$ & $60.03_{(1.15)}$ & $60.69_{(1.24)}$ & $62.23_{(1.70)}$ & $61.47_{(0.79)}$\\
        sohu-sts-B-sl & $65.56_{(1.69)}$ & $64.51_{(1.08)}$ & $68.08_{(3.01)}$ & $68.76_{(3.09)}$ & $70.34_{(0.84)}$\\
        sohu-sts-B-ss & $77.61_{(1.82)}$ & $80.05_{(0.86)}$ & $79.64_{(0.80)}$ & $80.03_{(0.97)}$ & $79.85_{(1.03)}$\\
        CBLUE-CHIP-STS & $75.80_{(1.21)}$ & $76.90_{(0.62)}$ & $75.91_{(1.12)}$ & $75.69_{(0.38)}$ & $77.90_{(0.59)}$\\
        CBLUE-KUAKE-QTR & $26.75_{(0.57)}$ & $27.00_{(0.56)}$ & $25.97_{(1.28)}$ & $26.11_{(0.77)}$ & $25.35_{(1.60)}$\\
        CBLUE-KUAKE-QQR & $43.57_{(2.03)}$ & $41.79_{(3.05)}$ & $38.47_{(7.19)}$ & $41.74_{(5.35)}$ & $35.35_{(8.27)}$\\
        PAWS-X & $53.52_{(0.64)}$ & $55.14_{(0.71)}$ & $54.19_{(0.59)}$ & $54.41_{(0.99)}$ & $54.90_{(0.37)}$\\
        nlpcc2016-dbqa & $63.89_{(2.07)}$ & $60.90_{(0.44)}$ & $64.24_{(2.68)}$ & $62.77_{(0.80)}$ & $62.61_{(3.64)}$\\
        cmrc2018\_public & $32.78_{(2.01)}$ & $33.24_{(2.70)}$ & $34.86_{(2.32)}$ & $32.07_{(1.51)}$ & $35.50_{(0.73)}$\\
        DRCD & $44.31_{(3.45)}$ & $43.08_{(2.69)}$ & $44.81_{(2.27)}$ & $43.11_{(1.91)}$ & $47.89_{(2.20)}$\\
        CCF2020-BDCI-QA & $13.05_{(1.13)}$ & $13.86_{(1.73)}$ & $15.27_{(0.91)}$ & $15.15_{(0.49)}$ & $16.22_{(0.56)}$\\
        CAIL2019-QA & $22.25_{(1.16)}$ & $21.31_{(1.11)}$ & $23.20_{(0.67)}$ & $20.61_{(1.48)}$ & $22.84_{(1.61)}$\\
        CAIL2020-QA & $27.90_{(1.48)}$ & $24.84_{(3.29)}$ & $26.45_{(1.50)}$ & $23.64_{(0.81)}$ & $26.87_{(2.14)}$\\
        msra\_ner & $57.18_{(4.84)}$ & $55.38_{(6.00)}$ & $57.88_{(5.04)}$ & $60.07_{(3.97)}$ & $58.17_{(4.40)}$\\
        weibo\_ner & $22.71_{(1.95)}$ & $23.24_{(0.95)}$ & $23.16_{(1.42)}$ & $23.28_{(1.62)}$ & $23.42_{(0.52)}$\\
        nlpcc2020-AutoIE & $33.65_{(6.85)}$ & $30.82_{(3.52)}$ & $33.95_{(3.15)}$ & $37.17_{(4.88)}$ & $35.29_{(6.25)}$\\
        CCF2020-BDCI-NER & $46.83_{(2.91)}$ & $45.45_{(3.76)}$ & $48.46_{(2.37)}$ & $47.35_{(3.30)}$ & $47.34_{(2.30)}$\\
        CMeEE & $24.87_{(3.15)}$ & $21.60_{(2.08)}$ & $25.59_{(3.58)}$ & $23.93_{(3.09)}$ & $24.84_{(0.94)}$\\
        SanWen-ner & $18.31_{(1.96)}$ & $16.72_{(1.79)}$ & $19.13_{(2.85)}$ & $17.82_{(1.96)}$ & $18.42_{(1.63)}$\\
        NLPCC2015 & $2.46_{(0.33)}$ & $2.47_{(0.47)}$ & $2.37_{(0.27)}$ & $2.45_{(0.46)}$ & $2.78_{(0.33)}$\\
        CAIL2020 & $0.86_{(0.16)}$ & $0.60_{(0.16)}$ & $0.82_{(0.32)}$ & $0.77_{(0.41)}$ & $0.81_{(0.05)}$\\
        WANFANG & $5.25_{(0.24)}$ & $5.23_{(0.81)}$ & $5.44_{(0.36)}$ & $5.46_{(0.42)}$ & $7.00_{(0.22)}$\\
        CSL\_SUMM & $1.48_{(0.22)}$ & $1.82_{(0.26)}$ & $1.74_{(0.47)}$ & $2.05_{(0.30)}$ & $3.35_{(0.55)}$\\
        EDU\_SUMM & $15.50_{(4.52)}$ & $14.74_{(1.89)}$ & $18.72_{(0.95)}$ & $15.04_{(2.67)}$ & $14.80_{(3.15)}$\\
        WEIBO & $4.95_{(0.94)}$ & $5.41_{(0.31)}$ & $4.95_{(0.67)}$ & $4.66_{(0.65)}$ & $5.45_{(0.45)}$\\
        COTE-BD & $6.81_{(1.61)}$ & $23.61_{(7.55)}$ & $20.79_{(3.38)}$ & $40.58_{(6.56)}$ & $48.29_{(9.36)}$\\
        COTE-MFW & $14.38_{(2.46)}$ & $32.34_{(9.76)}$ & $25.14_{(4.61)}$ & $43.81_{(6.53)}$ & $50.34_{(9.01)}$\\
        COTE-DP & $7.94_{(3.72)}$ & $18.46_{(9.97)}$ & $21.07_{(4.50)}$ & $23.89_{(10.29)}$ & $42.50_{(6.43)}$\\
        cluewsc2020\_public & $45.66_{(2.39)}$ & $42.76_{(1.40)}$ & $40.26_{(1.97)}$ & $42.06_{(1.35)}$ & $47.98_{(4.18)}$\\
        iflytek\_public & $18.99_{(2.70)}$ & $18.22_{(2.51)}$ & $23.95_{(3.17)}$ & $23.45_{(3.49)}$ & $26.14_{(1.02)}$\\
    \bottomrule
    \end{tabular}
    \caption{Detailed ablation results on prompt design and MLM loss}
    \label{tab:Detail_Table_for_Ablation_Studies}
\end{table*}

\begin{table*}[t]
\footnotesize
    \centering
    \begin{tabular}{lccccc}
    \toprule
         & \tabincell{c}{none} &
        \tabincell{c}{weighted avg \\all samples}   & 
        \tabincell{c}{weighted avg \\per sample}   & 
        \tabincell{c}{top1 \\ per sample}  & 
        \tabincell{c}{random init} \\ \midrule
        ALL & $44.83_{(0.55)}$ & $45.76_{(0.42)}$ & $46.01_{(0.52)}$ & $46.06_{(0.55)}$ & $46.16_{(0.54)}$\\ \midrule
        online\_shopping\_10cats & $95.49_{(0.30)}$ & $95.73_{(0.27)}$ & $95.73_{(0.27)}$ & $95.73_{(0.27)}$ & $95.72_{(0.27)}$\\
        ChnSentiCorp\_htl\_all & $92.92_{(0.51)}$ & $93.51_{(0.37)}$ & $93.42_{(0.37)}$ & $93.43_{(0.35)}$ & $93.45_{(0.38)}$\\
        nlpcc2014\_task2 & $79.90_{(0.29)}$ & $80.14_{(0.24)}$ & $80.14_{(0.23)}$ & $80.13_{(0.24)}$ & $80.12_{(0.24)}$\\
        yf\_dianping & $44.80_{(4.49)}$ & $44.63_{(4.68)}$ & $44.66_{(4.65)}$ & $44.63_{(4.66)}$ & $44.87_{(4.48)}$\\
        car\_sentiment & $24.44_{(1.81)}$ & $25.74_{(3.38)}$ & $25.73_{(3.37)}$ & $25.79_{(3.37)}$ & $25.80_{(3.41)}$\\
        dmsc & $38.21_{(2.38)}$ & $37.77_{(2.48)}$ & $37.81_{(2.30)}$ & $37.90_{(2.27)}$ & $37.88_{(2.31)}$\\
        weibo\_senti\_100k & $85.21_{(1.31)}$ & $85.45_{(0.94)}$ & $85.95_{(1.22)}$ & $85.91_{(1.23)}$ & $85.89_{(1.22)}$\\
        simplifyweibo\_4 & $39.54_{(3.07)}$ & $38.01_{(1.78)}$ & $38.67_{(1.76)}$ & $38.78_{(1.79)}$ & $38.87_{(2.06)}$\\
        NLPCC2014\_Weibo\_Emotion\_classification & $40.41_{(1.06)}$ & $41.23_{(1.18)}$ & $41.19_{(0.87)}$ & $41.22_{(0.94)}$ & $41.21_{(1.08)}$\\
        nCoV\_100k & $34.46_{(1.51)}$ & $34.86_{(1.32)}$ & $34.80_{(1.34)}$ & $34.82_{(1.38)}$ & $34.82_{(1.35)}$\\
        Internet\_News & $55.32_{(8.07)}$ & $55.12_{(8.58)}$ & $55.10_{(8.55)}$ & $55.19_{(8.58)}$ & $55.20_{(8.58)}$\\
        BDCI2019 & $35.69_{(5.31)}$ & $36.29_{(5.45)}$ & $36.46_{(5.43)}$ & $36.52_{(5.42)}$ & $36.53_{(5.45)}$\\
        SMP2019\_ECISA & $37.63_{(2.15)}$ & $38.49_{(1.90)}$ & $38.51_{(1.88)}$ & $38.51_{(1.87)}$ & $38.44_{(1.87)}$\\
        THUCNews & $65.58_{(3.27)}$ & $65.90_{(2.91)}$ & $65.89_{(2.91)}$ & $65.87_{(2.91)}$ & $65.86_{(2.93)}$\\
        CCFBDCI2020 & $75.61_{(4.08)}$ & $75.98_{(3.87)}$ & $75.86_{(4.13)}$ & $75.83_{(4.20)}$ & $75.93_{(4.21)}$\\
        tnews\_public & $46.04_{(1.26)}$ & $46.42_{(1.38)}$ & $46.36_{(1.42)}$ & $46.32_{(1.42)}$ & $46.35_{(1.50)}$\\
        Ifeng & $63.66_{(1.44)}$ & $62.78_{(1.20)}$ & $62.77_{(1.21)}$ & $62.77_{(1.18)}$ & $62.79_{(1.21)}$\\
        nlpcc2017\_news\_headline\_categorization & $46.95_{(1.36)}$ & $47.15_{(1.27)}$ & $47.16_{(1.31)}$ & $47.14_{(1.29)}$ & $47.14_{(1.37)}$\\
        catslu\_traindev & $90.55_{(0.74)}$ & $91.52_{(0.39)}$ & $91.57_{(0.42)}$ & $91.52_{(0.39)}$ & $91.33_{(0.54)}$\\
        e2e\_dials & $88.24_{(5.05)}$ & $86.38_{(5.55)}$ & $86.36_{(5.50)}$ & $86.44_{(5.53)}$ & $86.39_{(5.50)}$\\
        intent\_classification & $32.04_{(3.89)}$ & $34.37_{(4.37)}$ & $34.34_{(4.39)}$ & $34.37_{(4.37)}$ & $34.37_{(4.38)}$\\
        ocnli\_public & $46.98_{(1.96)}$ & $47.34_{(1.99)}$ & $47.21_{(2.06)}$ & $47.17_{(2.01)}$ & $47.16_{(2.09)}$\\
        afqmc\_public & $62.96_{(0.92)}$ & $63.51_{(0.87)}$ & $63.50_{(0.86)}$ & $63.50_{(0.86)}$ & $63.52_{(0.88)}$\\
        phoenix\_pair & $97.92_{(0.98)}$ & $98.99_{(0.20)}$ & $98.98_{(0.20)}$ & $98.99_{(0.20)}$ & $98.99_{(0.17)}$\\
        sohu-sts-A-ll & $64.97_{(0.57)}$ & $65.47_{(0.72)}$ & $65.47_{(0.73)}$ & $65.46_{(0.72)}$ & $65.44_{(0.72)}$\\
        sohu-sts-A-ss & $70.19_{(0.89)}$ & $70.80_{(0.67)}$ & $70.73_{(0.70)}$ & $70.72_{(0.74)}$ & $70.70_{(0.74)}$\\
        sohu-sts-B-ll & $61.81_{(1.39)}$ & $62.23_{(1.64)}$ & $62.22_{(1.61)}$ & $62.22_{(1.64)}$ & $62.23_{(1.70)}$\\
        sohu-sts-B-sl & $68.48_{(2.57)}$ & $68.77_{(3.11)}$ & $68.77_{(3.11)}$ & $68.76_{(3.11)}$ & $68.76_{(3.09)}$\\
        sohu-sts-B-ss & $79.77_{(0.78)}$ & $80.00_{(0.99)}$ & $79.99_{(0.94)}$ & $80.01_{(0.96)}$ & $80.03_{(0.97)}$\\
        CBLUE-CHIP-STS & $74.93_{(0.51)}$ & $75.66_{(0.36)}$ & $75.67_{(0.36)}$ & $75.67_{(0.36)}$ & $75.69_{(0.38)}$\\
        CBLUE-KUAKE-QTR & $25.73_{(0.85)}$ & $26.11_{(0.85)}$ & $26.14_{(0.86)}$ & $26.12_{(0.84)}$ & $26.11_{(0.77)}$\\
        CBLUE-KUAKE-QQR & $41.09_{(6.06)}$ & $41.62_{(5.20)}$ & $41.70_{(5.22)}$ & $41.62_{(5.21)}$ & $41.74_{(5.35)}$\\
        PAWS-X & $54.48_{(1.11)}$ & $54.39_{(0.96)}$ & $54.40_{(0.96)}$ & $54.39_{(0.96)}$ & $54.41_{(0.99)}$\\
        nlpcc2016-dbqa & $59.45_{(2.65)}$ & $62.86_{(0.87)}$ & $62.81_{(0.93)}$ & $62.84_{(0.87)}$ & $62.77_{(0.80)}$\\
        cmrc2018\_public & $34.43_{(1.64)}$ & $32.00_{(1.54)}$ & $31.94_{(1.54)}$ & $31.90_{(1.54)}$ & $32.07_{(1.51)}$\\
        DRCD & $42.99_{(3.90)}$ & $42.48_{(2.52)}$ & $42.57_{(2.50)}$ & $42.50_{(2.50)}$ & $43.11_{(1.91)}$\\
        CCF2020-BDCI-QA & $16.20_{(1.02)}$ & $14.96_{(0.53)}$ & $14.99_{(0.54)}$ & $15.15_{(0.69)}$ & $15.15_{(0.49)}$\\
        CAIL2019-QA & $20.88_{(2.19)}$ & $20.29_{(1.32)}$ & $20.52_{(1.47)}$ & $20.58_{(1.54)}$ & $20.61_{(1.48)}$\\
        CAIL2020-QA & $22.62_{(2.14)}$ & $23.29_{(0.84)}$ & $23.43_{(0.61)}$ & $23.61_{(0.63)}$ & $23.64_{(0.81)}$\\
        msra\_ner & $60.67_{(4.12)}$ & $60.05_{(4.45)}$ & $60.08_{(4.30)}$ & $60.00_{(4.13)}$ & $60.07_{(3.97)}$\\
        weibo\_ner & $23.20_{(1.60)}$ & $23.36_{(1.72)}$ & $23.47_{(1.80)}$ & $23.48_{(1.72)}$ & $23.28_{(1.62)}$\\
        nlpcc2020-AutoIE & $38.95_{(6.31)}$ & $35.92_{(4.59)}$ & $36.88_{(4.98)}$ & $36.78_{(4.95)}$ & $37.17_{(4.88)}$\\
        CCF2020-BDCI-NER & $47.51_{(4.18)}$ & $47.28_{(3.68)}$ & $47.35_{(3.40)}$ & $47.47_{(3.31)}$ & $47.35_{(3.30)}$\\
        CMeEE & $21.25_{(2.78)}$ & $24.26_{(3.27)}$ & $24.18_{(3.23)}$ & $23.80_{(3.11)}$ & $23.93_{(3.09)}$\\
        SanWen-ner & $18.26_{(1.91)}$ & $17.80_{(2.06)}$ & $17.85_{(2.03)}$ & $17.90_{(1.93)}$ & $17.82_{(1.96)}$\\
        NLPCC2015 & $2.05_{(0.33)}$ & $2.41_{(0.42)}$ & $2.37_{(0.44)}$ & $2.55_{(0.44)}$ & $2.45_{(0.46)}$\\
        CAIL2020 & $0.79_{(0.39)}$ & $0.74_{(0.42)}$ & $0.77_{(0.42)}$ & $0.81_{(0.45)}$ & $0.77_{(0.41)}$\\
        WANFANG & $5.64_{(0.52)}$ & $5.30_{(0.38)}$ & $5.32_{(0.32)}$ & $5.39_{(0.47)}$ & $5.46_{(0.42)}$\\
        CSL\_SUMM & $1.69_{(0.37)}$ & $1.89_{(0.25)}$ & $1.84_{(0.24)}$ & $1.91_{(0.33)}$ & $2.05_{(0.30)}$\\
        EDU\_SUMM & $16.81_{(1.73)}$ & $13.71_{(2.73)}$ & $14.80_{(2.94)}$ & $15.10_{(2.87)}$ & $15.04_{(2.67)}$\\
        WEIBO & $5.40_{(0.88)}$ & $4.61_{(0.62)}$ & $4.63_{(0.62)}$ & $4.68_{(0.65)}$ & $4.66_{(0.65)}$\\
        COTE-BD & $14.62_{(4.81)}$ & $26.80_{(4.97)}$ & $38.13_{(6.50)}$ & $39.09_{(7.09)}$ & $40.58_{(6.56)}$\\
        COTE-MFW & $16.35_{(5.31)}$ & $41.65_{(8.03)}$ & $40.64_{(7.40)}$ & $41.65_{(7.63)}$ & $43.81_{(6.53)}$\\
        COTE-DP & $12.21_{(7.17)}$ & $22.62_{(10.85)}$ & $22.69_{(10.79)}$ & $22.80_{(11.12)}$ & $23.89_{(10.29)}$\\
        cluewsc2020\_public & $43.11_{(0.63)}$ & $42.50_{(1.41)}$ & $42.50_{(1.41)}$ & $42.50_{(1.41)}$ & $42.06_{(1.35)}$\\
        iflytek\_public & $23.61_{(3.30)}$ & $23.39_{(3.50)}$ & $23.39_{(3.51)}$ & $23.37_{(3.41)}$ & $23.45_{(3.49)}$\\
        
    \bottomrule
    \end{tabular}
    \caption{Detailed ablation results on building new task-specific soft prompts}
    \label{tab:Detail_Table_for_Ablation_Studies_on_building_new_task_prompt_embedding}
\end{table*}

\end{document}